\documentclass[twoside]{article}

\usepackage[accepted]{aistats2023}
\usepackage[round]{natbib}

\usepackage{amsfonts}       
\usepackage{nicefrac}       
\usepackage{microtype}      
\usepackage{xcolor}         
\usepackage{colortbl}
\usepackage{bbm}
\usepackage{graphicx,setspace,latexsym,amsmath,amssymb,amsthm,bm}
\usepackage[font=small,labelfont=bf]{caption}
\usepackage{subcaption}
\usepackage{import}
\usepackage{dirtytalk}
\usepackage{algorithm,algorithmic}
\usepackage{subcaption}
\usepackage{changepage}
\usepackage{enumitem}
\usepackage[affil-it]{authblk}  
\usepackage{etoolbox}
\usepackage{lmodern}
\usepackage{tikz}
\usepackage[draft]{commands}
\usepackage{xr}
\usepackage{listings}
\usepackage[utf8]{inputenc} 
\usepackage[T1]{fontenc}

\DeclareFixedFont{\ttb}{T1}{txtt}{bx}{n}{8} 
\DeclareFixedFont{\ttm}{T1}{txtt}{m}{n}{8}  

\usepackage{color}
\definecolor{deepblue}{rgb}{0,0,0.5}
\definecolor{deepred}{rgb}{0.6,0,0}
\definecolor{deepgreen}{rgb}{0,0.5,0}

\usepackage{listings}

\newcommand\pythonstyle{\lstset{
language=Python,
basicstyle=\ttm,
morekeywords={self},              
keywordstyle=\ttb\color{deepblue},
emph={MyClass,__init__},          
emphstyle=\ttb\color{deepred},    
stringstyle=\color{deepgreen},
frame=tb,                         
showstringspaces=false
}}

\lstnewenvironment{python}[1][]
{
\pythonstyle
\lstset{#1}
}
{}

\newcommand\pythoninline[1]{{\pythonstyle\lstinline!#1!}}

\begin{document}

%

%

\twocolumn[

\aistatstitle{Optimal and Private Learning from Human Response Data}
\aistatsauthor{ Duc Nguyen \And Anderson Y. Zhang }
\aistatsaddress{ Department of Computer \& Information Science \\ University of Pennsylvania \And  Department of Statistics \& Data Science \\ University of Pennsylvania } ]

\begin{abstract} Item response theory (IRT) is the study of how people make probabilistic decisions, with diverse applications in education testing, recommendation systems, among others. The Rasch model of \emph{binary response data}, one of the most fundamental models in IRT, remains an active area of research with important practical significance. Recently, Nguyen and Zhang (2022) proposed a new spectral estimation algorithm that is efficient and accurate. In this work, we extend their results in two important ways. 

Firstly, we obtain a refined \emph{entrywise error bound} for the spectral algorithm, complementing the `average error' $\ell_2$ bound in their work. Notably, under mild sampling conditions, the spectral algorithm achieves the minimax optimal error bound (modulo a log factor). Building on the refined analysis, we also show that the spectral algorithm enjoys optimal sample complexity for top-$K$ recovery (e.g., identifying the best $K$ items from approval/disapproval response data), explaining the empirical findings in the previous work. 

Our second contribution addresses an important but understudied topic in IRT: privacy. Despite the human-centric applications of IRT, there has not been any proposed privacy-preserving mechanism in the literature. We develop a private extension of the spectral algorithm, leveraging its unique Markov chain formulation and the discrete Gaussian mechanism (Canonne et al., 2020). Experiments show that our approach is significantly more accurate than the baselines in the low-to-moderate privacy regime.
\end{abstract}

\section{INTRODUCTION}

Item response theory (IRT) was originally developed within the psychometric community in the early 60s \citep{rasch1960studies,van1997handbook} as a tool to model patients' responses to psychological tests. Since its conception, item response theory has been applied to diverse settings including education testing \citep{lord2012applications}, crowdsourcing \citep{whitehill2009whose}, recommendation systems \citep{chen2005personalized}, finance \citep{schellhorn2013using}, marketing \citep{brzezinska2016latent} and more recently the evaluation of machine learning algorithms \citep{moraes2020item,chen2020item,martinez2019item}.

One of the most fundamental models in IRT is the Rasch model \citep{rasch1960studies}. It assumes that the \emph{binary response $X_{li} \in \{0,1\}$} of person $l$ with latent characteristic $\theta^*_l \in \R$ to item $i$ with latent parameter $\beta^*_i\in \R$ is given as follows.
\begin{equation}\label{eqn:rasch-prob}
 \Pr(X_{li} = 1) = \frac{1}{1+\exp{-(\theta^*_l - \beta^*_i)}} \,.
\end{equation}
As an example in education testing, $\theta^*_l$ corresponds to the ability of student $l$ and $\beta^*_i$ the difficulty of problem $i$. The random response variable $X_{li}$ is whether the student correctly solves the problem.
Recently, \citet{nguyen2022spectral} proposed a new spectral algorithm for \emph{one-sided} parameter estimation under the Rasch model (i.e., estimate $\beta^*$). Experimentally, the spectral algorithm is significantly faster than its competitors while being comparatively accurate. As the spectral algorithm is entirely new, we identify two important open directions from the previous work.

\textbf{Entrywise Error Guarantee.} The guarantee in \citet{nguyen2022spectral} is an $\ell_2$ error bound for parameter estimation (cf., Theorem 3.3), which can be roughly considered the average estimation error. In certain domains, we might desire a small {entrywise error guarantee}. For example, consider a conference reviewing system where reviewers vote to reject or accept papers. We would like to accurately identify the top papers from reject/accept data. One approach is to estimate the quality parameters of the papers and accept $K$ papers with the higest parameter values\footnote{To adapt the Rasch model to this application, define $X_{li}=0$ if reviewer $l$ accepts paper $i$. $\beta^*_i$ corresponds to the quality of paper $i$ and $\theta^*_l$ the `harshness' of reviewer $l$.}. The $\ell_2$ error does not indicate how the estimation error distributes among the papers. The average error may still be low if all the error is concentrated in a single estimate. Really good papers could get rejected or subpar papers get accepted. On the other hand, a small entrywise parameter estimation error ensures that the quality for each paper is accurately measured.

\textbf{Privacy.} The study of privacy-preserving mechanisms in IRT is understudied. This is quite concerning because many item response theory models and algorithms are deployed on human response data containing sensitive information (e.g., a student passing a test, a voter supporting a legislation). One may consider the binary nature of the response data and propose randomized response \citep{warner1965randomized} as a natural solution. However, as we will show in our experiments, even when the number of responses per person is moderate $m \approx 10$, the amount of noise that randomized response needs to inject to ensure sufficient privacy is too large for any downstream algorithm to obtain accurate parameter estimates. On the other hand, analyzing the sensitivity (cf. Definition 3.1 in \citet{dwork2014algorithmic}) of the algorithms used in the IRT literature including optimization based algorithms \citep{andersen1973conditional,fischer1981existence,haberman1977maximum,hambleton1991fundamentals} is substantially more difficult. This might explain the seeming lack of progress towards developing a private algorithm in IRT.

\textbf{Our Contributions.} In this work, we make progress in both of the above open directions.
\vspace{-0.25cm}
\begin{itemize}
\item In section \ref{sect:entrywise-error}, we obtain refined entrywise analysis of the spectral algorithm, coupled with a matching lower bound, showing that under mild sampling conditions, the spectral algorithm achieves optimal entrywise error guarantee (up to a log factor). As a result of this optimality, we also show that the spectral algorithm can accurately identify all of the top $K$ items using only a constant factor more than the information theoretically minimal sample size. 
\item In Section \ref{sect:privacy}, we propose a privacy-preserving mechanism that is designed with the discrete nature of human response data in mind. We take advantage of the Markov chain formulation underlying the spectral algorithm and the discrete Gaussian mechanism to develop a conceptually simple yet performant private algorithm. The algorithm provides strong and controllable privacy guarantee and outperforms other approaches, especially in the low-to-moderate privacy regime.
\end{itemize}

\section{PROBLEM DESCRIPTION \& THE SPECTRAL ALGORITHM}\label{sect:problem}

We follow the notations of \cite{nguyen2022spectral}. Consider a universe with $n$ people and $m$ items. Each person $l$ has parameter $\theta^*_l \in \R$ and each item $i$ has parameter $\beta^*_i \in \R$. To overcome a fundamental identifiability issue associated with parameter translation, we assume a normalization constraint on the item parameters ${\beta^*}^\top \mb 1_m = 0$. We also assume that $\beta^*_i \in [\beta^*_{\min}, \beta^*_{\max}] \,\forall i\in[m]$ for some constants $\beta^*_{\min}$, $\beta^*_{\max}$. Similarly, we assume that $\theta^*_l \in [\theta^*_{\min}, \theta^*_{\max}]$ for some constants $\theta^*_{\min}, \theta^*_{\max}$.  Let $A \in \{0,1\}^{n\times m} $ denote the assignment matrix where $A_{li} = 1$ if person $l$ responds to item $i$ and $0$ indicates missing data. The observed data is $X \in \{0, 1, *\}^{n\times m}$ where $*$ denotes missing data. For entries where $A_{li} = 1$, $X_{li}$ is independently distributed per Equation (\ref{eqn:rasch-prob}). Let us consider the uniform sampling model where $A_{li} = 1$ with probability $p$ for some $p > 0$. 

\textbf{The Spectral Algorithm.} We summarize the original formulation of the spectral algorithm here. For more details on its implementation and analysis, we refer the reader to \citet{nguyen2022spectral}. For each item pair $i \neq j$, define the \emph{pairwise differential measurement} as
\begin{equation}\label{eqn:Yij}
Y_{ij} =  \sum_{l=1}^n A_{li} A_{lj} X_{li}(1-X_{lj}) \quad \forall i\neq j\in [m]\,.
\end{equation}
Note that this is a discrete-valued quantity. Given the pairwise differential measurements, the algorithm constructs a Markov chain $\hat M \in [0,1]^{m\times m}$ whose transition probabilities are defined as follows:
\begin{equation}\label{eqn:emp-markov-chain-def}
\hat M_{ij} = \begin{cases} \frac{1}{d} Y_{ij} &\text{ if $i \neq j$}\\
1 - \sum_{k\neq i} \frac{1}{d} Y_{ik} &\text{ if $i = j$}
\end{cases}\quad ,
\end{equation}
where $d$ is a sufficiently large normalization factor chosen so that the resulting pairwise transition probability matrix $\hat M$ does not contain any negative entries. Typically, $d = O(\max_{i\in [m]} \sum_{k\neq i} B_{ik})$ where $B_{ik} = \sum_{l=1}^n A_{li}A_{lk}$. The algorithm then computes the stationary distribution $\pi$ of $\hat M$ and recovers $\beta$ using a post-processing step. Algorithm \ref{alg:spectral} summarizes the spectral algorithm.
\begin{algorithm}[]
\caption{The Spectral Algorithm}
\hspace*{\algorithmicindent} \textbf{Input: } Binary response data $X\in \{0, 1, *\}^{n\times m}$. \\
\hspace*{\algorithmicindent} \textbf{Output: } Item parameter estimate $(\beta_i)_{i=1}^m$.\\
\vspace{-0.75em}
\begin{algorithmic}[1]
    \STATE Construct a Markov chain $\hat M$ per Equation (\ref{eqn:emp-markov-chain-def}).
    \STATE Compute the stationary distribution of $\hat{M}$:\\
           \quad Initialize $\pi^{(0)} = [\frac{1}{m},\ldots, \frac{1}{m}].$\\
           \quad For $t = 1, 2, \ldots$ until convergence, compute\\
           \quad $${\pi^{(t)}}^\top = \frac{{\pi^{(t-1)}}^\top  \hat M }{\lVert {\pi^{(t-1)}}^\top  \hat M \rVert_1 } \,.$$
    \STATE Compute $z = (\pi/d)/ \lVert \pi/d\rVert_1$ and $\bar\beta_i = \log z_i $ for $i \in [m]$. Return the normalized item parameters, i.e.,  $\beta = \bar\beta - \bar\beta^\top\mb 1/m$.\\
\end{algorithmic}
\label{alg:spectral}
\end{algorithm}



\section{FINE-GRAIN ANALYSIS OF THE SPECTRAL ALGORITHM}\label{sect:entrywise-error}

\subsection{Entrywise Error Bound}

In this section we provide refined entrywise error bound for the spectral algorithm. Recall the example with conference reviewing and that our practical motivation is to show that the estimation error incurred by the spectral algorithm `spreads out' among the parameters. This is to prevent the case where the estimation error is concentrated on a few items. 

\begin{theorem}\label{thm:entrywise-error} Consider the uniform sampling model described in Section \ref{sect:problem}. There exist constants $C_1, C_2$ such that if $np^2 \geq C_1 \log m$ and $mp \geq C_2 \log n$ then the output of the spectral algorithm satisfies
$$ \lVert \beta - \beta^* \rVert_{\infty} \leq \frac{C\sqrt{\log m}}{\sqrt{np}} $$
with probability at least $1- \bigO\left(m^{-10}\right) -\bigO\left(n^{-10}\right)$ where $C$ is an absolute constant.
\end{theorem}

The above theorem is a refinement of Theorem 3.3 in \citet{nguyen2022spectral}. In that paper, the authors show that the output of the spectral algorithm satisfies $ \lVert \beta - \beta^* \rVert_2 = \bigO\left( \frac{\sqrt{m}}{\sqrt{np}} \right) \,.$
Comparing the this $\ell_2$ bound with Theorem \ref{thm:entrywise-error} reveals that the magnitude of the entrywise error is no more than $\bigO\left(\sqrt{\frac{\log m}{m}}\right)$ the size of the $\ell_2$ error. This shows that the cummulative error indeed distributes evenly among the parameters and no two error terms differ in magnitude by more than a $\log$ factor.

Readers who are familiar with the learning to rank literature might see the parallel between this result and that in the Bradley-Terry-Luce (BTL) model where a similar algorithm -- spectral ranking -- enjoys a favorable entrywise error guarantee. However, the results are only superficially familiar and are based on similar fundamental theorems of Markov chain perturbation. Our results for the entrywise error bound and top-$K$ recovery in the next section are not simple extensions of the known results in the BTL literature. Firstly, the sampling model considered here fundamentally differs from the Erdos-Reyni comparison graph sampling model considered in the BTL model analysis. Additionally, the construction of the spectral algorithm is also different from the spectral ranking algorithm. Specifically, the pairwise transition probabilities and the construction of the Markov chain in Algorithm \ref{alg:spectral} and the spectral ranking algorithm are different.

Besides the spectral algorithm, we are only aware of entryise error guarantee for joint maximum likelihood estimate (JMLE) \citep{andersen1973conditional,haberman1977maximum}, formally stated in Theorem 2 of \citet{chen2021note}. However, it is well known that JMLE produces an inconsistent estimate of $\beta^*$ when $m$ is constant \citep{andersen1973conditional,ghosh1995inconsistent}. In the supplementary materials, we provide a complementary entrywise error bound for the spectral algorithm where we remove the assumption $mp = \Omega\left(\log n\right)$ from Theorem \ref{thm:entrywise-error}, showing that the spectral method always gives a consistent estimation of individual item parameters. On a practical note, the spectral algorithm has been shown to be similar in terms of accuracy but is significantly faster than JMLE \citep{nguyen2022spectral}.

\subsection{Top-$K$ Recovery Guarantee}
In this section, we analyze the performance of the spectral algorithm in terms of top-$K$ recovery, building on the entrywise error bound obtained earlier. Curiously, \citet{nguyen2022spectral} show through experimental results that the spectral algorithm often outperforms the baseline IRT algorithms in terms of top-$K$ accuracy. In this section, we provide a theoretical justification for its strong performance by furnishing an upper bound on the sample complexity for top-$K$ recovery of the spectral algorithm. Under mild sampling conditions, this upper bound turns out to have a matching lower bound which we will show in Section \ref{sect:lower-bound}.

To ground our discussion, we first define a theoretical quantity that captures the model-dependent difficulty of the top-$K$ recovery problem. 
Define $\Delta_K = \beta^*_{[K]} - \beta^*_{[K+1]}$ to be the gap between the $K$-th and the $K+1$-th best items. 
The smaller the gap, the harder it is to separate the top $K$ items from the remaining items. 

The reader could intuitively see that if we have a finite-sample entrywise error bound on $\beta$, we can obtain a finite sample error guarantee for top-$K$ recovery. Suppose that we have a sufficiently large sample size such that $ \lVert \beta - \beta^* \rVert_\infty \leq \frac{\Delta_K}{2} \,. $
Then we separate all of the top $K$ items from the bottom $n-K$ items and achieve perfect top $K$ recovery.
Building on this intuition, we have the following top-$K$ recovery guarantee for the spectral algorithm.
\begin{theorem}\label{thm:top-K} Consider the setting of Theorem \ref{thm:entrywise-error}. Consider a top-$K$ estimator that first runs the spectral algorithm on the response data and returns the $K$ items with the highest parameter values. There exists a constant $C_K$ such that if $np \geq \frac{C_K\log m}{\Delta_K^2} $, then this top-$K$ estimator correctly identifies all of the top $K$ items
with probability at least $1- \bigO\left(m^{-10}\right) - \bigO\left(n^{-10}\right)$.
\end{theorem}

To summarize, the above theorem states that so long as $m, n, p$ satisfy
$$ np = \Omega\left(\frac{\log m}{\Delta_K^2}\right), np^2 =\Omega\left(\log m\right) \,\text{and}\, mp = \Omega\left(\log n\right) \,,$$
then the spectral method accurately identifies all of the top $K$ items. To the best of our knowledge, this guarantee (and the lower bound shown in the next section) is the first formal result for the ranking performance of any  estimation algorithm in the literature.

\subsection{Lower Bounds}
In this section, we complement the upper bounds obtained in previous sections with corresponding information theoretic lower bounds.

Firstly, \textbf{for parameter estimation}, we obtain a lower bound result that generalizes Theorem 3.5 of \citet{nguyen2022spectral}. Note that Theorem 3.5 in the previous work is a Cramer-Rao bound that only applies to \emph{unbiased estimators}. However, our \emph{new} information theoretic lower bound is strictly more general and is applicable to \emph{all statistical estimators}.
\begin{theorem}\label{thm:lower-bound-error} Consider the sampling model described in Section \ref{sect:problem} and further assume that $np = \Omega(1)$. There exists a class of Rasch models $\B$ such that for any statistical estimator, the minimax risk is lower bounded as
$$  \inf_{\hat \beta}\sup_{\beta^* \in \B}\, \E \lVert\hat \beta - \beta^* \rVert_2^2 = \Omega\left(\frac{m}{np}\right)\,. $$
\end{theorem}

As a consequence of the normed inequality $\lVert \beta - \beta^* \rVert_{\infty} \geq \frac{1}{\sqrt{m}} \lVert \beta - \beta^* \rVert_2$, we can lower bound the minimax entrywise error for any statistical estimator as
$$ \inf_{\hat \beta}\sup_{\beta^*\in \B}\, \E  \lVert\hat \beta - \beta^* \rVert_{\infty}^2  = \Omega\left(\frac{1}{np}\right)\,. $$
From Theorem \ref{thm:entrywise-error}, the spectral algorithm satisfies $\lVert \beta - \beta^* \rVert^2_{\infty} = \bigO\left(\frac{\log m}{{np}}\right) \,. $
This directly establishes the optimality (modulo a log factor) of the spectral algorithm.

Moving on to \textbf{top-$K$ recovery guarantee}, the following theorem establishes the minimum sample complexity required to accurately identify all of the top $K$ items from user-item binary response data.
\begin{theorem}\label{thm:lower-bound-topK} Consider the sampling model described in Section \ref{sect:problem}. There is a class of Rasch model such that if 
$np \leq \frac{c_K\log m}{\Delta_K^2} \,,$
where $c_K$ is a constant, then any estimator will fail to identify all of the top $K$ items with probability at least $\frac{1}{2}$.
\end{theorem}
We compare the above lower bound to the upper bound obtained in the previous section. The condition $mp = \Omega\left(\log n\right)$ is a mild one, requiring that the number of items grows slowly and is appropriate in applications such as recommendation systems or crowdsourcing.
We see that $np = \Omega\left(\frac{\log m}{\Delta_K^2} \right)$ is the necessary sample size to accurate identify all of the top-$K$ items and the spectral algorithm needs the same condition.
The spectral algorithm also requires $np^2 = \bigO\left(\log m\right)$. This is a consequence of the design of the algorithm that operates on pairwise differential measurements. 

In general, $\Delta_K$ may be arbitrarily small and change with $m$. For example, if each $\beta^*_i$ is uniformly sampled from $[\beta_{\min}^*, \beta_{\max}^*]$ then $\E[\Delta_K] = \bigO(\frac{1}{m})$.
This makes comparisons between the two conditions on $n$ difficult. In some real life situations, however, one can assume that $p = \bigO(1)$. This is appropriate in settings such as education testing where each student is shown a constant fraction of all questions from a question bank, or in recommendation systems where the users are shown a constant fraction of a product catalogue. In this regime of $p$, the lower bound simplifies to $n  = \Omega\left(\frac{\log m}{\Delta_K^2} \right)$. Assuming that $\Delta_K = \smallO(1)$, the spectral algorithm needs $n =\bigO\left( \max\left\{\frac{\log m}{\Delta_K^2}, \log m \right\} \right) = \bigO\left( \frac{\log m}{\Delta_K^2} \right) $ to accurately identify all of the top $K$ items -- the optimal sample complexity for top-$K$ recovery.

\vspace{-0.25cm}\label{sect:lower-bound}


\section{THE PRIVATE SPECTRAL ALGORITHM}\label{sect:privacy}
\subsection{A Brief Introduction to Differential Privacy}
In this section, we propose, for the first time in the IRT literature, a privacy-preserving estimation algorithm. We focus on differential privacy (DP), which has become the de facto framework for analyzing the privacy guarantees of machine learning algorithms. To simplify the description, we consider the \emph{full response setting} -- a person responds to all items. The starting point is the definition of two neighboring datasets.
\begin{definition} Let $X, X'$ be two binary responses data as described in Section \ref{sect:problem}. We write $X \simeq X'$ to denote that $X$ and $X'$ are neighboring datasets if $X$ and $X'$ differ by the entries of exactly one row.
\end{definition}
Note that we are trying to provide \emph{user privacy} in our application. Intuitively, a randomized function or a statistical query of a dataset (e.g., counting the number of people who respond positively to an item) provides privacy to the users if its output is not sensitive to changes in a single user's data. This is formalized by the definition (approximate) differential privacy.
\begin{definition} [Approximate Differential Privacy \citep{dwork2014algorithmic}] For any input domain $\X$ and output domain $\Y$, a randomized function $\tilde f: \X \rightarrow \Y$ satisfies $(\epsilon, \delta)$-differential privacy if for any two $X \simeq X'$ and any $C \subseteq  \Y$, $ \Pr(\tilde f(X')\in C) \leq e^{\epsilon} \, \Pr(\tilde f(X) \in C) + \delta \,. $
\end{definition}
The quantity $\epsilon$ is the \emph{privacy budget} or desired privacy level. The smaller its value, the stronger the privacy guarantee but the private output tends to be less accurate. When an algorithm makes multiple private functions evaluation or queries on the same dataset, we call it \emph{query composition}. In general, privacy guarantee \emph{decays linearly} with the number of queries \citep{dwork2014algorithmic,dwork2016concentrated}. That is, for a composition of $k$ queries to satisfy overall $\epsilon^*$ privacy level, the privacy budget for each query is approximately $\epsilon^*/k$. When $k$ is large, the accuracy of DP algorithms tend to suffer significantly. Motivated by this practical consideration, researchers have developed refined notions of privacy that enjoys better accuracy under query composition than DP. One of these is concentrated differential privacy (CDP).
\begin{definition} [Concentrated Differential Privacy \citep{dwork2016concentrated,bun2016concentrated}] For any input domain $\X$ and output domain $\Y$, a randomized function $\tilde f: \X \rightarrow \Y$ satisfies $\frac{\epsilon^2}{2}$- concentrated differential privacy if for all $X \simeq X'$ and $\alpha \in (1,\infty)$,
$ D_{\alpha}(\tilde f(X) \,\lVert\, \tilde f(X') ) \leq \frac{\epsilon^2 \alpha}{2} \,,$
where $D_{\alpha}(P \lVert Q) = \frac{1}{\alpha - 1} \log \sum_{y\i \Y} P(y)^{\alpha}Q(y)^{1-\alpha} $ is the R\'{e}nyi divergence between the distributions $P$ and $Q$.
\end{definition}

\subsection{The Discrete Gaussian Mechanism}
A common approach to preserve privacy is to add random noise to the output of non-private functions. Generally, adding a larger the amount of noise gives better the privacy guarantee at the expense of downstream estimation accuracy. 
Within this class of mechanisms, the discrete Gaussian mechanism \citep{canonne2020discrete} is a recently proposed noise adding mechanism utilizing a discretized generalization of the Gaussian distribution.
We deliberately choose a discrete noise mechanism because it has been found that even innocuous finite-precision calculations can be exploited by malicious users to recover the original discrete data exactly \citep{mironov2012significance}. Given the discrete nature of the binary response data, we believe that this approach is pertinent.

Secondly, we prefer the discrete Gaussian mechanism to the discrete Laplace mechanism which adds discrete Laplace noise \citep{ghosh2009universally}. The discrete Gaussian mechanism has over the Laplace mechanism is that the Gaussian distribution has lighter tail than the Laplace distribution. As our experiments will show, when we have to repeatedly make a large number of queries, the total variance of the Gaussian noise is lower than that of the Laplace noise needed to achieve a comparable privacy level and the resulting estimate is more accurate.

For the design of our private spectral algorithm, recall the definition of a pairwise differential measurement in Equation (\ref{eqn:Yij}). For a fixed privacy budget $\epsilon$, define the \emph{privatized pairwise differential measurement} as
\begin{equation}\label{eqn:private-Yij}
\tilde Y_{ij} =  \sum_{l=1}^n A_{li} A_{lj} X_{li}(1-X_{lj}) + Z_{ij} \quad \forall i\neq j\in [m]\,,
\end{equation}
where each $Z_{ij} \sim \calN_{\bbZ}(0, \frac{1}{\epsilon^2})$ is an independent random variable drawn from the 0-mean discrete Gaussian distribution with variance $\frac{1}{\epsilon^2}$ (see Algorithm 1 of \cite{canonne2020discrete}). 

There are $m(m-1)$ pairwise differential measurements. Hence, a privatized spectral algorithm that adds $\calN_{\bbZ}(0, \frac{1}{\epsilon^2})$ to each pairwise differential measurement $Y_{ij}$ achieves $\frac{1}{2}m(m-1)\epsilon^2$-CDP overall \citep{canonne2020discrete}. Conversely, if we desire an overall privacy level $\epsilon^*$, we can compute the noise variance needed for each pairwise differential measurement. 

\begin{lemma}\label{lem:privacy-guarantee} Consider a modified spectral algorithm that first adds $\calN_{\bbZ}\left(0, \frac{\sqrt{m(m-1)}}{{\epsilon^*}^2}\right)$-distributed noise to each pairwise differential measurement per Equation (\ref{eqn:private-Yij}). The algorithm then proceeds to use the private differential measurements to construct the Markov chain as in the original algorithm. Then the modified spectral algorithm satisfies $\frac{1}{2}{{\epsilon^*}^2}$-CDP.
\end{lemma}

\subsection{Changing the Privacy-Accuracy Tradeoff via Sparsification}

From Lemma \ref{lem:privacy-guarantee}, one can see that the variance of the additive Gaussian noise scales as the \emph{square root of the number of pairwise differential measurements}. When $n \gg m$, the additive noise will be dominated by the signal from the user data and the private spectral algorithm will produce an accurate estimate (e.g., see Figure \ref{fig:synthetic}).
On the other hand, suppose we have a large $m$ (say $m\geq 100$) but a relatively small $n$ ($n \leq 500$). Adding $\calN_{\bbZ}\left(0, \frac{\sqrt{m(m-1)}}{\epsilon^2}\right)$-distributed noise to each pairwise differential measurement may introduce a lot of noise, dominating the signal and leading to a less accurate final estimate. 

To overcome this limitation, we propose a heuristic modification to the private spectral algorithm. Instead of constructing a ``dense'' Markov chain where every pairwise transition probabilities are non-zero, we construct a \emph{sparse approximation} of this Markov chain. We first generate an Erd\H{o}s-R\'{e}nyi graph with $m$ vertices and edge probability $p_0 = (1+\smallO(1)) \frac{\log m}{m}$. Note that the condition on the edge probability is to ensure that the resulting comparison graph is connected with high probability \citep{erdHos1960evolution}. If we further ensure that all pairwise transition probabilities are positive\footnote{This can be achieved by adding regularization as described in \citet{nguyen2022spectral}}, the resulting Markov chain will admit a unique stationary distribution and subsequently a unique parameter estimate. The advantage here is that we only compute $(1+\smallO(1)) m\log m$ pairwise differential measurements. Hence for each subsampled pairwise differential measurement, we only need to add $\calN_{\bbZ}\left(0, \frac{\sqrt{\bigO(m\log m)}}{{\epsilon^*}^2}\right)$-distributed noise and achieve the same CDP guarantee, a $\sqrt{m}$ factor reduction in variance compared to the dense Markvo chain construction.

However, sparsification might also come with its own limitation. By removing some of the pairwise transition probabilities information from the Markov chain, the final estimate obtained from the sparse Markov chain may deviate from that using full information. As our experiments will show, sparsification tradeoffs between additive noise and approximation error. It is appropriate to use sparsification to achieve a low-to-moderate privacy guarantee with good estimation accuracy when the number of items is large and the number of user is small. Finally, Algorithm \ref{alg:spectral-private} summarizes our private spectral algorithm. 

\begin{algorithm}[]
\caption{The Private Spectral Estimator}
\hspace*{\algorithmicindent} \textbf{Input: } Response data $X\in \{0, 1\}^{n\times m}$, $p_0$ (default 1) and desired privacy level $\epsilon^*$.  \\
\hspace*{\algorithmicindent} \textbf{Output: } Private item estimate $\tilde \beta$\\
\vspace{-1em}
\begin{algorithmic}[1]
    \STATE Construct a private Markov chain $\tilde M$ with underlying comparison graph drawn from an Erdos-Reyni distribution with edge probability $p_0$ per Equation (\ref{eqn:private-Yij}) and desired privacy level $\epsilon^*$.
    \STATE Follow steps 2-3 in Algorithm \ref{alg:spectral} using $\tilde M$.
\end{algorithmic}
\label{alg:spectral-private}
\end{algorithm}


\section{EXPERIMENTS}
In this section, we evaluate the privacy-accuracy tradeoff of the private spectral algorithm on synthetic and real-life datasets. We use randomized response (on the user data) and the discrete Laplace mechanism as baseline methods. Recently it has been shown that one can amplify the privacy guarantee of randomized response by shuffling the randomized user data (cf., Theorem III.1 of \citet{feldman2022hiding}). We compare the algorithms in terms of their approximate differential privacy guarantee \footnote{One can convert concentrated differential privacy guarantee of the spectral algorithm to approximate differential privacy guarantee using Lemma 3.5 of \citet{bun2016concentrated}.}. Specifically, we fix a common $\delta = 10^{-4}$ and evaluate the accuracy at varying privacy levels $\epsilon^*$. For readers who are not familiar with privacy-preserving mechanisms, we describe these approaches in the supplementary materials and their implementation.

\subsection{Synthetic Datasets.} 
\begin{figure}[htp]
\centering
\begin{subfigure}{.35\textwidth}
  \centering
  \includegraphics[width=1.\linewidth]{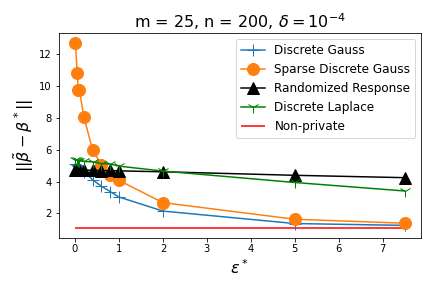}
  \label{fig:sfig2}
\end{subfigure}\\
\vspace{-0.2cm}
\begin{subfigure}{.35\textwidth}
  \centering
  \includegraphics[width=1.\linewidth]{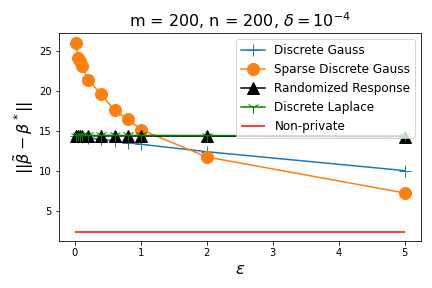}
 \caption{Given the same number of users, increasing $m$ expectedly leads to a worse privacy-accuracy tradeoff for all methods. The discrete Gaussian mechanism generally outperforms the baselines at the low-to-moderate privacy regime.}
  \label{fig:sfig2}
\end{subfigure}\\
\begin{subfigure}{.35\textwidth}
  \centering
  \includegraphics[width=1.\linewidth]{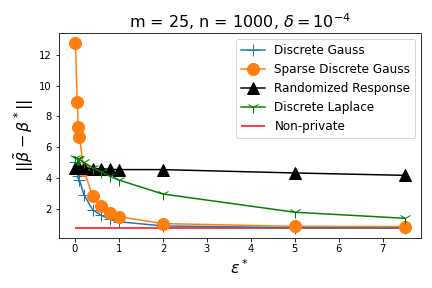}
  \label{fig:sfig3}
\end{subfigure}\\
\vspace{-0.2cm}
\begin{subfigure}{.35\textwidth}
  \centering
  \includegraphics[width=1.\linewidth]{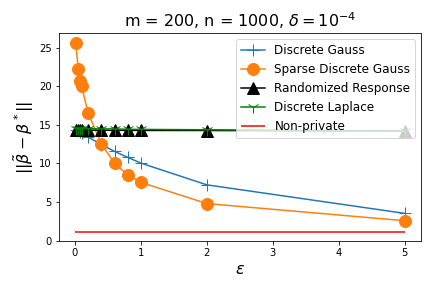}
  \caption{The privacy-accuracy tradeoff for the discrete Gaussian mechanism gets better as $n$ increases (from 200 in the previous subfigure to 1000) while randomized response and the Laplace mechanism see little gain from a larger $n$.}
  \label{fig:sfig4}
\end{subfigure}
\caption{ \textbf{Synthetic datasets.} The discrete Gaussian mechanism outperforms the baselines at low-to-moderate privacy regime and enjoys substantial improvement from a larger sample size $n$.}
\label{fig:synthetic}
\end{figure}

Figure \ref{fig:synthetic} shows the performance of the algorithms on synthetic datasets where we know the ground truth $\beta^*$. The amount of noise introduced by the discrete Gaussian mechanism gets proportionally smaller than the signal as $n$ grows. Expectedly, given a sufficiently large $n$ (relative to $m$), we see almost identical performance between the non-private estimator and the private estimator using the discrete Gaussian mechanism. At the very high privacy regime $\epsilon^* \approx 0.01$, all methods produce very inaccurate results. Sparsification is significantly worse than the other approaches in this regime because of the combined effect of a large amount of additive noise and the approximation error introduced by sparsification. The differences among the methods become more apparent in the low-to-moderate privacy regime. As $\epsilon^*$ increases, the accuracy of the discrete Gaussian mechanism improves significantly. On the other hand, randomized response and the Laplace mechanism would require either a very large $n$ or a very large privacy budget $\epsilon^*$ to produce a reasonably accurate estimate, neither of which is desirable. 

\subsection{Real-life Datasets.} For real life datasets, we do not have a ground truth $\beta^*$. However, previous work has shown that the spectral algorithm is competitive with the most commonly used IRT algorithms. Also considering that we focus on privacy-preservation in this work, we evaluate $\lVert \tilde \beta - \beta\rVert$, where $\beta$ is the output of the non-private spectral algorithm and $\tilde\beta$ that of the corresponding private algorithm.

\textbf{Education Datasets.} We include commonly used education testing datasets including LSAT \citep{mcdonald2013test}, UCI \citep{hussain2018educational} and the Three Grades dataset \citep{cortez2008using}. These datasets have a very small number of items. At very high privacy level $\epsilon \approx 0.01$, all methods produce very inaccurate estimate and randomized response is the most accurate method. However, at a moderate privacy level $\epsilon^* \approx 1$, the Laplace mechanism and the Gaussian mechanism outperform the randomized response approach. On the other hand, to produce a reasonably accurate estimate, randomized response would require a very large $\epsilon^*$, leading to a vacuous privacy guarantee.
\begin{figure}[htp]
\vspace{-0.2cm}
\centering
\begin{subfigure}{.35\textwidth}
  \centering
  \includegraphics[width=1.\linewidth]{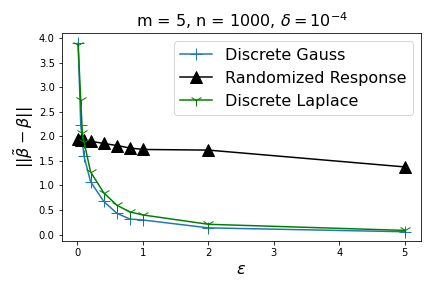}
  \caption{\textbf{LSAT dataset.}}
  \label{fig:sfig2}
\end{subfigure}\\
\vspace{-0.1cm}
\begin{subfigure}{.35\textwidth}
  \centering
  \includegraphics[width=1.\linewidth]{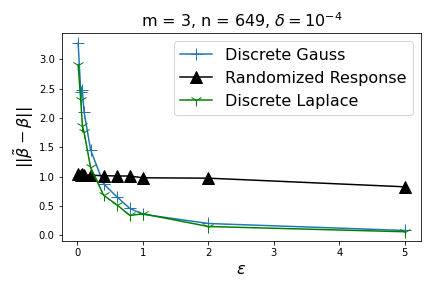}
  \caption{\textbf{Three grades dataset.}}
  \label{fig:sfig3}
\end{subfigure}
\vspace{-0.1cm}
\begin{subfigure}{.35\textwidth}
  \centering
  \includegraphics[width=1.\linewidth]{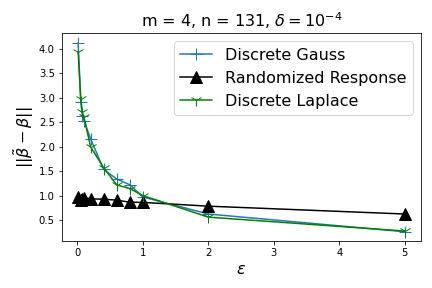}
  \caption{\textbf{UCI student dataset.}}
  \label{fig:sfig4}
\end{subfigure}
\caption{\textbf{Education datasets.} The Gaussian mechanism and the Laplace mechanism outperforms randomized response at low-to-moderate privacy levels. The results here suggest that for many education testing application where the number of items is small, the spectral algorithm with the discrete Gaussian (or Laplace) mechanism should be the preferred privacy-preserving algorithm.}
\label{fig:education}
\end{figure}

\begin{figure}[htp]
\centering
\begin{subfigure}{.4\textwidth}
  \centering
  \includegraphics[width=1.\linewidth]{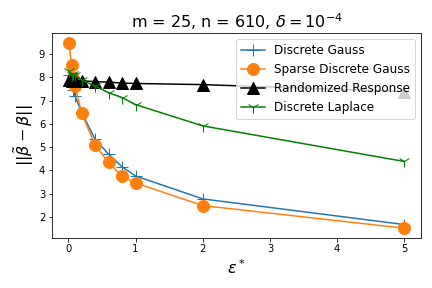}
  \caption{When $m$ is small, sparsification doesn't significantly improve the performance of the private spectral algorithm.}
  \label{fig:sfig2}
\end{subfigure}\\
\begin{subfigure}{.4\textwidth}
  \centering
  \includegraphics[width=1.\linewidth]{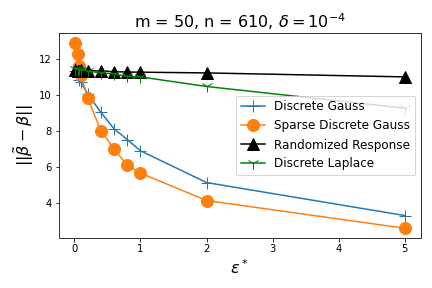}
  \label{fig:sfig2}
\end{subfigure}\\
\vspace{-0.25cm}
\begin{subfigure}{.4\textwidth}
  \centering
  \includegraphics[width=1.\linewidth]{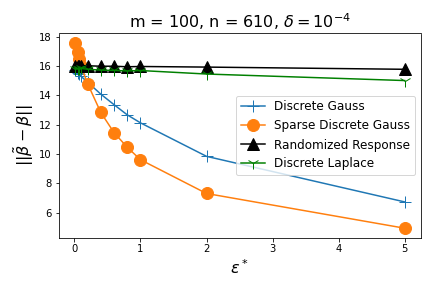}
  \label{fig:sfig3}
\end{subfigure}\\
\vspace{-0.25cm}
\begin{subfigure}{0.4\textwidth}
  \centering
  \includegraphics[width=1.\linewidth]{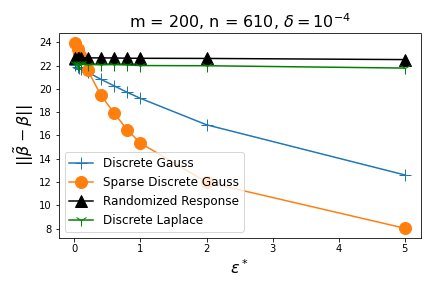}
  \caption{With the same $n$ and a large number of items $m$, the `sparse Markov chain' approach is significantly better than the `dense Markov chain' approach.}
  \label{fig:sfig3}
\end{subfigure}
\caption{\textbf{ML-100K datasets.} The discrete Gaussian mechanism generally outperforms the baselines, especially in the low-to-moderate privacy regime $\epsilon^* \geq 1$.}
\label{fig:recsys}
\end{figure}

\textbf{ML-100K Recommendation Systems Datasets.} For experiments on recommendation systems datasets \citep{harper2015movielens}, we follow a similar approach to previous work to generate binary response data from ratings data \citep{lan2018estimation, davenport20141}. We first perform a low rank matrix estimation of the user-item ratings matrix and use the estimation as the user-item preference scores. For each user, we convert scores that are lower than the average to 1 and scores that are higher than the average to 0. We then select a random subset of $m$ items and extract the corresponding submatrix as the response matrix. From Figure \ref{fig:recsys}, one can see that the discrete Gaussian mechanism provides superi
or privacy-accuracy tradeoff compared to the other approaches. Notably, sparsification provides a more accurate parameter estimation by reducing the amount of noise needed to ensure the same level of privacy and this improvement is significant when $m$ is large. 

From the experiments on synthetic and real-life datasets, one can see that when we desire high accuracy and a moderate level of privacy, the discrete Gaussian mechanism should be the preferred approach. Additionally, when the number of items $m$ is large while the number of users $n$ is small, one should use the sparse Markov chain construction.

\section{ETHICAL DISCUSSIONS}

Our work explores privacy matters for an algorithm of which real-life applications often involve human data with sensitive information. For example, the Rasch model is often studied in the context of education testing and psychological testing where the subjects are students and patients. Therefore, deploying any algorithms in such capacity needs to be accompanied by thoughtful and thorough ethical considerations. In this work, we provide the algorithmic tool to quantify the error associated with the parameter estimates and to preserve the privacy of the human subjects. 

\section{CONCLUSION}
In this work, we made two contributions to the understanding of the spectral algorithm for parameter estimation under the Rasch model. Firstly, we show that the spectral algorithm achieves the optimal entrywise error bound, complementing its optimal $\ell_2$ error guarantee obtained in the previous work. Secondly, we propose, for the first time in the item response theory literature, a differentially private algorithm. We complement our theoretical and algorithmic contributions with experimental results showing that the discrete Gaussian mechanism provides accurate parameter estimation, outperforming the baselines. A possible future extension of this work is a deeper analysis of the random subsampling of the comparison graph in the sparse discrete Gaussian mechanism. We believe that the randomized sparsification step could provide some form of privacy amplification. Furthermore, designing an accurate estimation algorithm for the high privacy regime ($\epsilon^* \approx 0.01$) remains an unresolved challenge.

\section{ACKNOWLEDGEMENT}
The authors would like to thank the anonymous reviewers for their thoughtful suggestions and comments. We thank Aaron Roth and Emily Diana of the University of Pennsylvania for fruitful discussions on the topic of differential privacy. The authors are supported by NSF Grant DMS-2112099. A.Z. acknowledges financial support from the Alfred H. Williams Faculty Scholar award. Any opinions expressed in this paper are those of the authors and do not necessarily reflect the views of the National Science Foundation.

\paragraph{Addressing the Reviewers' Feedback.} In response to the anonymous reviewers' feedback during the submission process, we have performed additional theoretical and experimental analyses and included the additional findings in the supplementary materials. Specifically, we have studied how the performance of the sparse spectral algorithm depends on the edge sampling probability $p_0$. We have also provided a high-level analysis of the entrywise error of the \emph{private} spectral estimate, complementing Theorem \ref{thm:entrywise-error}.

\bibliography{reference}

\end{document}


\makeatletter\@input{extra.tex}\makeatother

\onecolumn

\maketitle

%

%



\tableofcontents


\section{Proofs of Entrywise Error Guarantee}
\subsection{Preliminaries}
Consider the following idealized Markov chain where the state transition probabilities are exact:
\begin{equation*}\label{eqn:ideal-mc}
 M^*_{ij} = \begin{cases} \frac{1}{d} Y_{ij}^* &\text{for } i\neq j \\ 1 - \frac{1}{d} \sum_{k\neq i} Y^*_{ik} &\text{for } i = j \end{cases} \quad ,
\end{equation*}
where $Y_{ij}^* =  \sum_{l=1}^n A_{li} A_{lj} \E[X_{li}(1-X_{lj})]$. Define $\pi^*_i = \frac{e^{\beta^*_i}}{\sum_{k=1}^m e^{\beta^*_k} }$ for $i \in [m]$. It is known that $\pi^*$ is the stationary distribution of the Markov chain $M^*$ \cite{nguyen2022spectral}. Define $\pi^*_{\max} := \max_{i\in [m]}\{ \pi_i^* \}$, $\pi^*_{\min} := \min_{i\in [m]}\{ \pi_i^* \}$. Define $ \kappa:= \beta^*_{\max} - \beta^*_{\min}$. One can show that $\pi^*_{\max}/\pi^*_{\min} \leq e^{\kappa}$. 

Define $\gamma := \min_{l\in[n], i, j\in [m]} \E[X_{li}(1-X_{lj})]$.
As a convenient notation, we use $a\gtrsim b$ to mean $a = \Omega(b)$ and $a\asymp b$ to mean $a \gtrsim b$ and $b \gtrsim a$. Define $a \lesssim b$ analogously. We use $a \lor b$ to mean $\max\{a, b\}$ and $a \land b$ to mean $\min \{a, b\}$. Define $B = A^\top A$ (i.e., $B_{ij} = \sum_{l=1}^n A_{li}A_{lj}$) and the following two events.
$$ \A = \{ \frac{np^2}{2} \leq B_{ij} \leq \frac{3np^2}{2},\forall i\neq j\in [m] \}\,,$$
$$ \A^+ = \A \cap \{ \frac{mp}{2} \leq A_{l}^\top \mb 1 \leq \frac{3mp}{2} \,\forall l \in [n] \} \,.$$
Conditioning on these two events simplify substantially the proofs of various upper bounds that the rest of the paper is devoted to. Fortunately, both events happen with high probability under reasonable sampling assumptions, as summarized by two lemmas below. 
\begin{lemma}\label{lem:epsilon_C_event} Consider the random sampling scheme described in Section \ref{sect:problem}. There exists a constant $C_0$ such that if $np^2 \geq C_0 \log m$ then
$$ \Pr(\A) \geq 1 -  2\,\exp{-\frac{np^2}{24}} $$
\end{lemma}
\begin{proof} Invoking Chernoff bound, we have for a pair $i\neq j$,
\begin{equation*}
\begin{aligned}
\Pr\left(\lvert \sum_{l=1}^n A_{li}A_{lj} - \E[A_{li}A_{lj}] \rvert > \frac{1}{2}np^2\right) \leq 2 \,\exp{-\frac{np^2 }{12}}\,.
\end{aligned}
\end{equation*}
Applying union bound over all pairs, we have
\begin{align*}
&\Pr\left(\exists i\neq j: \,\lvert \sum_{l=1}^n A_{li}A_{lj} - \E[A_{li}A_{lj}] \rvert > \frac{1}{2}np^2 \,  \right)\\
&\leq {m\choose 2} \cdot 2\,\exp{-\frac{np^2}{12}} \leq 2 \exp{-\frac{np^2}{12} + 2\log m} \\
&\leq 2\,\exp{-\frac{np^2}{24}}\,,
\end{align*}
where the last inequality holds so long as $np^2 \geq 48\log m$.
\end{proof}

\begin{lemma} Consider the random sampling scheme described in Section \ref{sect:problem}. There exist constants $C_0$ and $C_0'$ such that if $np^2 \geq C_0 \log m$ and $mp \geq C_0' \log n$, then
$$ \Pr(\A^+) \geq 1 - 2\,\exp{-\frac{np^2}{24}} - \frac{2}{n^{10}}\,. $$
\end{lemma}
\begin{proof} The first term is obtained using the same argument as in the proof of Lemma \ref{lem:epsilon_C_event}. For the second term, we again invoke Chernoff bound.
\begin{equation*}
\begin{aligned}
\Pr(\lvert \sum_{l=1}^n A_{li}A_{lj} - \E[A_{li}A_{lj}] \rvert > \frac{1}{2}mp) \leq 2\,\exp{-\frac{\frac{1}{4}mp }{\frac{1}{2}+2}}
\leq 2\, \exp{-\frac{mp}{12}}\,.
\end{aligned}
\end{equation*}
Applying union bound over all users $n$ gives
\begin{align*}
&\Pr\left(\exists l\in[n]: \,  \lvert \sum_{l=1}^n A_{li}A_{lj} - \E[A_{li}A_{lj}] \rvert > \frac{1}{2}mp \,\right) \\
&\leq n \cdot 2\,  \exp{-\frac{mp}{10}+\ln 2}\\
&= 2\, \exp{-\frac{mp}{10}+ \log n}\\
&\leq \frac{2}{n^{10}}\,,
\end{align*}
where the last inequality holds so long as $mp \geq 110 \log n$.
\end{proof}

\emph{Conditioned on $\A$ (or $\A^+$)}, one can set $d = \frac{3mnp^2}{2}$ to ensure that the Markov chain $M$ constructed in Algorithm \ref{alg:spectral} contains no negative entries. For the remaining of the proof, we will explicitly make use of this choice. Additionally, one can verify the following useful inequalities.
\begin{align*}
\pi^*_{\min} &\geq \frac{1}{me^{\kappa}}\,.\\
\pi^*_{\max} &\leq \frac{e^{\kappa}}{m}\,. 
\end{align*}

\subsection{Basic Entrywise Analysis}
To begin our entrywise analysis, pick any index $i \in [m]$. We have the following deterministic decomposition.
\begin{equation*}
\begin{aligned}
\pi_i - \pi_i^* &= (\pi^\top M)_i - ({\pi^*}^\top M^* )_i\\
&= \sum_j \pi_j M_{ji} - \sum_j \pi_j^* M_{ji}^*\\
&= \sum_j \pi_j M_{ji} - \sum_j \pi_j^*(M_{ji}^* - M_{ji} + M_{ji})\\
&= \sum_j (\pi_j - \pi_j^*)M_{ji} - \sum_j \pi_j^*(M^*_{ji} - M_{ji})\\
&= \sum_j (\pi_j - \pi_j^*)M_{ji} + \sum_j \pi_j^*(M_{ji} - M^*_{ji})\,.\\
&= \underbrace{\sum_{j\neq i} (\pi_j - \pi_j^*)M_{ji}}_{I_1^{(i)}} + (\pi_i - \pi_i^*)M_{ii} + \underbrace{\sum_{j\neq i} \pi_j^*(M_{ji} - M^*_{ji})}_{I_3^{(i)}} + \underbrace{\pi_i^*(M_{ii}-M_{ii}^*)}_{I_4^{(i)}}\,.
\end{aligned}
\end{equation*}
Rearranging the terms gives
\begin{equation}\label{eqn:entrywise-decomp}
\pi_i - \pi_i^* = \frac{1}{\underbrace{1-M_{ii}}_{I_2^{(i)}}  } \cdot \bigg(\underbrace{\sum_{j\neq i} (\pi_j - \pi_j^*)M_{ji}}_{I_1^{(i)}} + \underbrace{\sum_{j\neq i} \pi_j^*(M_{ji} - M^*_{ji})}_{I_3^{(i)}} + \underbrace{\pi_i^*(M_{ii}-M_{ii}^*)}_{I_4^{(i)}} \bigg) \,.
\end{equation}

The goal in this section and the next is to provide a bound on the absolute deviation of the RHS by bounding each of the terms $\pow{I_1}{(i)},\pow{I_2}{(i)},\pow{I_3}{(i)},\pow{I_4}{(i)} $ separately. The main difference between the analysis in this section versus that in the next section is the added assumption that $m$ grows with $\log n$, $mp \gtrsim \log n$. In this section, we do not assume anything about $m$. The goal here is to show that when $m$ is a constant or grows very slowly with $n$, the spectral algorithm produces consistent entrywise estimate of the parameter $\beta^*$. This is in sharp contrast to JMLE which produces an inconsistent estimate of $\beta^*$ under this regime \citep{andersen1973conditional,ghosh1995inconsistent}. In the next section, we obtain better (in fact optimal) entrywise error bound when $m$ grows at least logarithmically with $n$.

We first establish that the empirical Markov chain $M$ is entrywise close to the true Markov chain $M^*$. 
\begin{proposition}\label{prop:calE} Suppose that $\A$ holds. There exists a constant $C_1$ such that the following event
\begin{equation} \label{ev:calE}
\calE = \bigg\{  \lvert M_{ij} - M_{ij}^* \rvert \leq \frac{C_1 \sqrt{\log np^2 \lor \log m} }{\sqrt \gamma\,\sqrt{np^2}} \cdot M_{ij}^* \,\,\,\forall i\neq j \bigg\} \,.
\end{equation}
happens with probability at least $1 - 2 \left( (np^2)^{-10} \land m^{-10} \right)$.
\end{proposition}

\begin{proof} For a single pair $(i, j)$, note that
$$ M_{ij} - M_{ij}^* = \frac{1}{d} \sum_{l=1}^n A_{li}A_{lj} \bigg(X_{li}(1-X_{lj}) - \E[X_{li}(1-X_{lj})]  \bigg) $$
is a sum of $n$ independent random variables. Using Chernoff bound, we have, for $\alpha \in (0,1)$,
\begin{align*}
&\Pr\bigg( \lvert M_{ij} - M^*_{ij} \rvert > \alpha \, M_{ij}^* \bigg\lvert \A \bigg)\\
&\leq 2\,\exp{-\frac{\alpha^2  {M_{ij}^*} }{3}} \leq 2 \,\exp{-\frac{\alpha^2 \left(B_{ij}\gamma \right) }{3}}\\
&\leq 2\,\exp{-\frac{\alpha^2\,\gamma\, np^2 }{12}}\,.
\end{align*}
Applying union bound over all pairs $i\neq j$ gives
\begin{align*}
&\Pr\bigg( \exists i \neq j \,:\, \lvert M_{ij} - M^*_{ij} \rvert > \alpha \, M_{ij}^* \bigg\lvert \A \bigg)\\
&\leq m^2 \,2\,\exp{-\frac{\alpha^2\,\gamma\, np^2 }{12}}\\
&\leq 2\,\exp{-\frac{\alpha^2\,\gamma\, np^2 }{12} + 2\log m}\\
&\leq 2\,\exp{-\frac{\alpha^2\,\gamma np^2}{24}}\,,
\end{align*}
where the last inequality holds so long as $\frac{\alpha^2 \gamma np^2}{24} \geq 2\log m$. Setting $\alpha = \frac{\sqrt{10}\sqrt{24} \sqrt{\log np^2 \lor \log m}}{\sqrt{\gamma} \sqrt{np^2}}$ is a valid choice and the constant $C_1$ in the proposition's statement is $\sqrt{240}$.
\end{proof}

We first show that, conditioned on $\A$ and $\calE$, we can obtain deterministic upper bound on $I_2^{(i)}, I_3^{(i)}$, $I_4^{(i)}$.

\begin{lemma}\label{lem:i3-i4-bound} Suppose that both $\A$ and $\calE$ hold and that $np^2 > \frac{C^2_1}{\gamma} \left(\log np^2 \lor \log m\right) $ where $C_1$ is the same constant in Proposition \ref{prop:calE}. Then the following inequalities hold deterministically for all $i \in [m]$.
$$ I_2^{(i)} \geq \bigg( 1- \frac{C_1 \sqrt{\log np^2 \lor \log m} }{\sqrt \gamma\,\sqrt{np^2}} \bigg) \cdot \frac{\gamma}{3}\,, $$ 
$$ \lvert I_3^{(i)} \rvert, \lvert I_4^{(i)}\rvert \leq \frac{C_1 \sqrt{\log np^2 \lor \log m} }{\sqrt \gamma\,\sqrt{np^2}} \cdot \lVert \pi^* \rVert_{\infty}  \,.$$
\end{lemma}

\begin{proof} When $\A$ holds, recall that $B_{ij} = \sum_{l} A_{li}A_{lj} \leq \frac{3np^2}{2}$ and we set $d = \frac{3mnp^2}{2}$. We have
\begin{align*}
1 - M_{ii} &= \sum_{j\neq i} M_{ij} \geq \sum_{j\neq i} \left(1-\frac{C_1 \sqrt{\log np^2 \lor \log m} }{\sqrt \gamma\,\sqrt{np^2}}\right) M_{ij}^*\\
&= \left(1-\frac{C_1 \sqrt{\log np^2 \lor \log m} }{\sqrt \gamma\,\sqrt{np^2}}\right) \cdot \sum_{j\neq i} \frac{1}{d} \, \sum_{l=1}^n A_{li}A_{lj} \E[X_{li}(1-X_{lj})]\\
&\geq \left(1-\frac{C_1 \sqrt{\log np^2 \lor \log m} }{\sqrt \gamma\,\sqrt{np^2}}\right) \cdot \frac{1}{2d} (m-1)np^2\gamma \\
&\leq \left(1-\frac{C_1 \sqrt{\log np^2 \lor \log m} }{\sqrt \gamma\,\sqrt{np^2}}\right) \cdot \frac{\gamma}{3} \,.
\end{align*}
We will now bound the term $I_3^{(i)}$.
\begin{align*}
&\sum_{j\neq i} \pi_j^* (M_{ji} - M_{ji}^*) \\
&\leq \frac{C_1 \sqrt{\log np^2 \lor \log m} }{\sqrt \gamma\,\sqrt{np^2}} \cdot \sum_{j\neq i} \pi_j^* M_{ji}^* =  \frac{C_1 \sqrt{\log np^2 \lor \log m} }{\sqrt \gamma\,\sqrt{np^2}}\cdot \pi_i^* \leq \frac{C_1 \sqrt{\log np^2 \lor \log m} }{\sqrt \gamma\,\sqrt{np^2}} \cdot \lVert \pi^* \rVert_{\infty} \,.
\end{align*}
~\\
Next, we bound $I_4^{(i)}$, using a similar argument as that used to bound $I_3^{(i)}$, we have
\begin{align*}
\pi_i^* (M_{ii} - M_{ii}^*) &= \pi_i^* \cdot \sum_{j\neq i} (M_{ij}^* - M_{ij})\\
&\leq \pi_i^* \cdot \sum_{j\neq i}  \frac{C_1 \sqrt{\log np^2 \lor \log m} }{\sqrt \gamma\,\sqrt{np^2}} \, M_{ij}^* =  \frac{C_1 \sqrt{\log np^2 \lor \log m} }{\sqrt \gamma\,\sqrt{np^2}}\, \cdot \sum_{j\neq i} \pi_i^* M_{ij}^*\\
&= \frac{C_1 \sqrt{\log np^2 \lor \log m} }{\sqrt \gamma\,\sqrt{np^2}}\, \cdot \sum_{j\neq i} \pi_j^* M_{ji}^* =  \frac{C_1 \sqrt{\log np^2 \lor \log m} }{\sqrt \gamma\,\sqrt{np^2}}\, \pi_i^*  \leq \frac{C_1 \sqrt{\log np^2 \lor \log m} }{\sqrt \gamma\,\sqrt{np^2}} \cdot \lVert \pi^* \rVert_{\infty}\,,
\end{align*}
where between the second and third line we use the fact that the Markov chain $M^*$ is time-reversible (Proposition 2.1 of \citet{nguyen2022spectral}). That is, $\pi_i^* M_{ij}^* = \pi_j^* M_{ji}^*$.
\end{proof}

~\\
What remains is to bound the term $I_1^{(i)}$. By Cauchy-Schwarz,
\begin{equation}\label{eqn:I1}
I_1^{(i)} \leq \lVert \pi - \pi^* \rVert_2 \cdot \sqrt{\sum_{j\neq i} M_{ji}^2 } \,. 
\end{equation}
\citet{nguyen2022spectral} have shown an upper bound on $\lVert \pi - \pi^*\rVert_2$. We modify their argument and present the result as follows.

\begin{lemma}\label{lem:pi-ell2-bound} Suppose that Condition $\A$ and Condition $\calE$ hold. There exists a constant $C_2$ such that if $np^2 \geq \frac{e^{4\kappa}C_2}{\gamma^3} \left(\log np^2 \lor \log m\right) $ then
\begin{equation*}
\lVert\pi-\pi^*\rVert_2 \leq \frac{1}{\frac{\gamma}{6e^{2\kappa}}}\,\frac{C_3\lVert \pi^* \rVert_{\infty} \sqrt{\log np^2 \lor m }}{\sqrt{np^2}}
\end{equation*}
with probability at least $1 - \exp{-40m} \land (np^2)^{-40}$, where $C_3$ is an absolute constant.
\end{lemma}

\begin{proof}
For the $\norm{\pi- \pi^*}$ term of (\ref{eqn:I1}), we invoke the $\ell_2$ analysis of the spectral method (cf. Lemma A.3 of \citet{nguyen2022spectral}), we have the following useful deterministic inequality
\begin{equation}\label{eqn:ssd-bound}
 \lVert \pi - \pi^* \rVert_2 \leq \frac{\lVert {\pi^*}^\top(M^* - M)\rVert_{2}  }{\mu(M^*) - \lVert M - M^* \rVert_{2} }\,,
 \end{equation}
where $\mu(M^*)$ is the spectral gap of $M^*$. From Lemma A.6 of \citet{nguyen2022spectral}, we have
\begin{equation}\label{eqn:lemA6}
 \mu(M^*) \geq \frac{\gamma}{3e^{2\kappa}}\,. 
 \end{equation}
Note that $M_{ij}^* \leq \frac{B_{ij}}{d} \leq \frac{1}{m}$ for any $i\neq j$. We have
\begin{equation*}
\begin{aligned}
\lVert M - M^* \rVert_{2} &\leq \lVert \diag(M) I - \diag(M^*) I\rVert_{2} + \lVert [M - M^*]_{i\neq j}\rVert_{2} \\
&\leq \max_{i} \lvert M_{ii} - M_{ii} \rvert + \max_{u, v: \lVert u \rVert = \lVert v\rVert = 1 } \sum_{i\neq j} u_i (M_{ij} - M_{ij}^*)v_j\\
&\leq \max_{i} \lvert \sum_{j\neq i} M_{ij}-M_{ij}^* \rvert + \max_{i\neq j} \lvert M_{ij}-M^*_{ij}\rvert \cdot \sum_{i\neq j} \lvert u_i\rvert \lvert v_j\rvert\\
&\leq 2 m \cdot \max_{i\neq j} \lvert M_{ij} - M_{ij}^* \rvert \leq 2m \,  \frac{C_1 \sqrt{\log np^2 \lor \log m} }{\sqrt \gamma\,\sqrt{np^2}} \, \max_{i\neq j} M^*_{ij} \leq \frac{2C_1 \sqrt{\log np^2 \lor \log m} }{\sqrt \gamma\,\sqrt{np^2}}\,.
\end{aligned}
\end{equation*}
So long as
\begin{equation}\label{eq:condition1}
np^2 \geq \frac{144e^{4\kappa}C_1^2 \left(\log np^2 \lor \log m\right)}{\gamma^3} 
\end{equation}
then $\frac{C_1\sqrt{\log np^2 \lor \log m}}{\sqrt{\gamma}\sqrt{np^2}} \leq \frac{\gamma}{12e^{2\kappa}}$ and the denominator of (\ref{eqn:lemA6}) is bounded as
$$ \mu(M^*) - \lVert M - M^* \rVert_{2} \geq \frac{\gamma}{3e^{2\kappa}} - \frac{2C_1 \sqrt{\log np^2 \lor \log m} }{\sqrt \gamma\,\sqrt{np^2}} \geq \frac{\gamma}{3e^{2\kappa}} - \frac{\gamma}{6e^{2\kappa}} = \frac{\gamma}{6 e^{2\kappa}}. $$ 
For the numerator term of (\ref{eqn:lemA6}), $\lVert {\pi^*}^\top(M^* - M)\rVert_{2}$, we follow a simlar argument as in the proof of Lemma A.9 of \citet{nguyen2022spectral}. We have, for any $C' > 1$,
\begin{align*} 
\Pr\bigg( \lVert {\pi^*}^\top (P - M^*)\rVert_2 &>  \frac{\lVert \pi^* \rVert_{\infty} \sqrt{4C' \max\{m, \log np^2\}  }}{\sqrt{3np^2}} \,\bigg\lvert \A \bigg) \leq \exp{-4(C'-1)\max\{m, \log np^2\} }\\
& = \min\left\{\frac{1}{\exp{4(C'-1)m}}, \frac{1}{(np^2)^{4(C'-1)}}\right\} 
\end{align*}
Set $C' = 11$, and substitute the bounds for $ \mu(M^*) - \lVert M - M^* \rVert_{2} \geq \frac{\gamma}{6e^{2\kappa}} $ into (\ref{eqn:ssd-bound}). This finishes the proof.
\end{proof}


What remains to bound $I_1^{(i)}$ is to obtain an upper bound on $\sqrt{\sum_{j\neq i} M_{ji}^2 }$ in (\ref{eqn:I1}).
\begin{lemma} \label{lem:I1-const} Suppose that the setting and conclusion of Lemma \ref{lem:pi-ell2-bound} hold. Then the following inequality holds simultaneously for all $i \in [m]$ deterministically.
$$\lvert I_1^{(i)} \rvert \leq \frac{1}{\sqrt{m}} \,\frac{1 + \frac{\gamma }{12e^{2\kappa}}}{\frac{\gamma}{6e^{2\kappa}}} \cdot \frac{C_3\lVert \pi^* \rVert_{\infty} \sqrt{\log np^2 \lor m }}{\sqrt{np^2}}\,. $$
\end{lemma}

\begin{proof} We have
\begin{align*}
\sqrt{\sum_{j\neq i} M_{ji}^2 } &\leq \sqrt{\sum_{j\neq i} \left(M_{ji}^* \left(\frac{1}{\sqrt{m}}  \frac{C_1 \sqrt{\log np^2 \lor \log m} }{\sqrt \gamma\,\sqrt{np^2}}\right)\right)^2 } \\
&\leq \left({1+ \frac{C_1 \sqrt{\log np^2 \lor \log m} }{\sqrt \gamma\,\sqrt{np^2}}}\right) \cdot \sqrt{\sum_{j\neq i} {M_{ji}^*}^2 } \\
&\leq \frac{1}{\sqrt{m}}+ \frac{C_1 \sqrt{\log np^2 \lor \log m} }{\sqrt \gamma\,\sqrt{mnp^2}}\\
&\leq \left(1 + \frac{\gamma}{12e^{2\kappa}}\right) \,\frac{1}{\sqrt m} \,.
\end{align*}
The third inequality comes from the fact that conditioned on $\A$ and $\calE$, $M_{ji}^* \leq \frac{B_{ij}}{d} \leq \frac{3/2np^2}{3/2mnp^2} = \frac{1}{m}$. The last inequality comes from the condition on $np^2$ in Lemma \ref{lem:pi-ell2-bound}. The final lemma statement is obtained by combining Lemma \ref{lem:pi-ell2-bound} and the upper bound obtained above into (\ref{eqn:I1}). 
\end{proof}


\begin{theorem}\label{thm:pi-error-const} Suppose that Condition $\A$ and $\calE$ hold. Suppose further that $np^2 > \frac{C^2_1}{\gamma} \left(\log np^2 \lor \log m\right) \lor \frac{e^{4\kappa}C_2}{\gamma^3} \left(\log np^2 \lor \log m\right) $ where $C_1, C_2$ are two absolute constants. Then
\begin{equation*}
\lVert {\pi - \pi^*}\rVert_{\infty} \leq \frac{1}{\frac{\gamma}{3}\left(1 -\frac{\gamma}{12e^{2\kappa}}\right) } \, \left(\frac{2C_1}{\sqrt{\gamma}}\, \frac{\sqrt{\log np^2 \lor \log m}}{\sqrt{np^2}} + \frac{1 + \frac{\gamma }{12e^{2\kappa}}}{\frac{\gamma}{6e^{2\kappa}}}  \, \frac{C_3\sqrt{\log np^2/m \lor 1 }}{\sqrt{np^2}} \right) \, \lVert\pi^*\rVert_{\infty}
\end{equation*}
with probability at least $1 - \exp{-40m} \land (np^2)^{-40}$.
\end{theorem}

\begin{proof} Conditioned on $\A$ and $\calE$, the bounds on $I_2^{(i)}, I_3^{(i)}, I_4^{(i)}$ per Lemma \ref{lem:i3-i4-bound} hold deterministically. Under the stated assumptions on $np^2$, the bound on $I_1^{(i)}$ per Lemma \ref{lem:I1-const} holds with probability at least $1 - \exp{-40m} \land (np^2)^{-40}$. Substituting the corresponding upper bounds into (\ref{eqn:entrywise-decomp}) completes the proof.
\end{proof}

~\\
The result we have obtained provides a bound on $\frac{\lVert \pi - \pi^* \rVert_{\infty}}{\lVert\pi^*\rVert_{\infty}}$. Recall that we are ultimately interested in $\lVert \beta - \beta^* \rVert_{\infty}$. 

~\\
\begin{theorem}\label{thm:beta-err-const} There exist absolute constants $C_0, C_1, C_2, C_4$ such that if $np^2 > C_0\log m \lor \frac{C^2_1}{\gamma} \left(\log np^2 \lor \log m\right) \lor \frac{e^{4\kappa}C_2}{\gamma^3} \left(\log np^2 \lor \log m\right) \lor \frac{C_4e^{6\kappa}}{\gamma^4} (\log np^2 \lor \log m) $ then
\begin{equation*}
\lVert \beta - \beta^* \rVert_{\infty} \leq\frac{4e^{2\kappa}}{\frac{\gamma}{3}\left(1 -\frac{\gamma}{12e^{2\kappa}}\right) } \, \left(\frac{2C_1}{\sqrt{\gamma}}\, \frac{\sqrt{\log np^2 \lor \log m}}{\sqrt{np^2}} + \frac{1 + \frac{\gamma }{12e^{2\kappa}}}{\frac{\gamma}{6e^{2\kappa}}}  \, \frac{C_3\sqrt{\log np^2/m \lor 1 }}{\sqrt{np^2}} \right)
\end{equation*}
with probability at least $1 - 2\,\exp{-\frac{np^2}{24}} - 2 \left((np^2)^{-10} \land m^{-10}\right) - \left(\exp{-40m} \land (np^2)^{-40}\right)$.
\end{theorem}

\begin{proof} Conditioned on $\A$ and $\calE$, by Theorem \ref{thm:pi-error-const} and the fact that $\lVert \pi^*\rVert_{\infty} \leq \frac{e^{\kappa}}{m}$,
$$\lVert {\pi - \pi^*}\rVert_{\infty} \leq \frac{1}{\frac{\gamma}{3}\left(1 -\frac{\gamma}{12e^{2\kappa}}\right) } \, \left(\frac{2C_1}{\sqrt{\gamma}}\, \frac{\sqrt{\log np^2 \lor \log m}}{\sqrt{np^2}} + \frac{1 + \frac{\gamma }{12e^{2\kappa}}}{\frac{\gamma}{6e^{2\kappa}}}  \, \frac{C_3\sqrt{\log np^2/m \lor 1 }}{\sqrt{np^2}} \right) \, \frac{e^{\kappa}}{m} \,$$
with probability at least $1 - \left(\exp{-40m} \land (np^2)^{-40}\right) $. One could verify that so long as both of the following conditions on $np^2$ hold
\begin{align*}
np^2 &\geq \frac{64C_1^2 e^{2\kappa}}{\gamma^3\left(1 - \frac{\gamma}{12e^{2\kappa}}\right)^2 } \left(\log np^2 \lor \log m\right)  \,,\\
np^2 &\geq \frac{72^2e^{6\kappa}}{\gamma^4}\,\left(\frac{1 + \frac{\gamma}{12e^{2\kappa}}}{1 - \frac{\gamma}{12e^{2\kappa}}}\right)^2 \log np^2\,,
\end{align*}
then
\begin{equation*}
\frac{1}{\frac{\gamma}{3}\left(1 -\frac{\gamma}{12e^{2\kappa}}\right) } \, \left(\frac{2C_1}{\sqrt{\gamma}}\, \frac{\sqrt{\log np^2 \lor \log m}}{\sqrt{np^2}} + \frac{1 + \frac{\gamma }{12e^{2\kappa}}}{\frac{\gamma}{6e^{2\kappa}}}  \, \frac{C_3\sqrt{\log np^2/m \lor 1 }}{\sqrt{np^2}} \right) \leq \frac{1}{2e^{\kappa}}
\end{equation*}
and $\lVert \pi - \pi^* \rVert_{\infty} \leq \frac{1}{2me^{\kappa}}$. Within the range $\left[\frac{1}{2me^{\kappa}}, \frac{e^{\kappa}}{m} + \frac{1}{2me^{\kappa}} \right]$, the log function satisfies $\lvert \log x - \log x' \rvert \leq 2me^{\kappa} \lvert x - x'\rvert$. The output of the spectral algorithm is
$$ \beta_i = \log\pi_i - \frac{1}{m} \sum_{k=1}^m \log \pi_k \,. $$
$\beta^*$ is similarly related to $\pi^*$:
$$ \beta^*_i = \log \pi^*_i - \frac{1}{m} \sum_{k=1}^m \log \pi^*_k \,. $$
Then for all $i \in [m]$,
\begin{equation}\label{eqn:beta-ell-inf}
\begin{aligned}
\lvert \beta_i - \beta^*_i \rvert &\leq \lvert \log\pi_i - \log\pi_i^* \rvert + \frac{1}{m} \sum_{k=1}^m \lvert \log \pi_k - \log \pi_k^* \rvert\\
&\leq 2 \max_{k} \, \lvert \log\pi_k - \log \pi_k^* \rvert \\
&\leq 4me^{\kappa} \, \max_{k} \, \lvert \pi_k - \pi_k^* \rvert\\
&\leq 4me^{\kappa} \cdot \lVert \pi - \pi^* \rVert_{\infty}\\
&\leq \frac{4e^{2\kappa}}{\frac{\gamma}{3}\left(1 -\frac{\gamma}{12e^{2\kappa}}\right) } \, \left(\frac{2C_1}{\sqrt{\gamma}}\, \frac{\sqrt{\log np^2 \lor \log m}}{\sqrt{np^2}} + \frac{1 + \frac{\gamma }{12e^{2\kappa}}}{\frac{\gamma}{6e^{2\kappa}}}  \, \frac{C_3\sqrt{\log np^2/m \lor 1 }}{\sqrt{np^2}} \right)\,.
\end{aligned}
\end{equation}
To obtain the probability bound shown in the theorem statement, we have
\begin{align*}
&\Pr\left( \lnot (\ref{eqn:beta-ell-inf}) \right) \leq \Pr\left( \lnot (\ref{eqn:beta-ell-inf}) , \A, \calE \right) + \Pr\left( \lnot (\ref{eqn:beta-ell-inf}) , \lnot \A \text{ or }  \lnot \calE  \right)\\
&\leq \Pr\left( \lnot (\ref{eqn:beta-ell-inf}) \,\bigg|\, \A, \calE \right)  \cdot \Pr\left(\A, \calE\right) + \Pr\left( \lnot (\ref{eqn:beta-ell-inf}) \,\bigg|\, \lnot \A \text{ or }  \lnot \calE  \right) \cdot \Pr\left(  \lnot \A \text{ or }  \lnot \calE\right)\\
&\leq \left(\exp{-40m} \land (np^2)^{-40}\right) + 2\,\exp{-\frac{np^2}{24}} + 2 \left((np^2)^{-10} \land m^{-10}\right).
\end{align*}
This completes the proof.
\end{proof}


\newpage
\subsection{Improved Rates for the Regime $mp\gtrsim \log n$}

In this section, we present the proof of Theorem \ref{thm:entrywise-error} in the main paper. The first lemma is analogous to Lemma \ref{lem:i3-i4-bound} but we obtain a tighter bound on $\abs{\pow{I_3}{(i)}}$ and $\abs{\pow{I_4}{(i)}}$.

\begin{lemma}\label{lem:I3-I4-growing} Suppose that both $\A^+$ and $\calE$ hold and that $np^2 > \frac{C_1^2}{\gamma}\left(\log np^2 \lor \log m\right)$ where $C_1$ is the same constant as in Proposition \ref{prop:calE}. Then the following inequality holds deterministically for all $i \in [m]$.
$$ I_2^{(i)} \geq \bigg( 1- \frac{C_1 \sqrt{\log np^2 \lor \log m} }{\sqrt\gamma\,\sqrt{np^2}} \bigg) \cdot \frac{\gamma}{3}\,. $$ 
On the other hand, there also exists a constant $C_5$ such that the following holds across all $i \in [m]$
$$ \lvert I_3^{(i)} \rvert, \lvert I_4^{(i)}\rvert \leq \frac{C_5\sqrt{\log m} }{\sqrt{np}} \cdot \lVert \pi^* \rVert_{\infty}  \,$$
with probability at least $1 - 4 m^{-10}$.
\end{lemma}

\begin{proof} Note that event $\A^+$ is a special case of event $\A$. Therefore, the bound on $\pow{I_2}{(i)}$ is still true by the proof of Lemma \ref{lem:i3-i4-bound}. 
It remains to obtain a sharper bound for $I_3^{(i)}$ and $I_4^{(i)}$. We note in the proof of Lemma \ref{lem:i3-i4-bound} that both of these two terms can be bounded in the same way. Therefore, it suffices to bound the term $I_3^{(i)}$ using the method of bounded difference. We have
\begin{equation*}
\begin{aligned}
I_3^{(i)} &= \sum_{j\neq i} \pi_j^* (P_{ji} - P_{ji}^*) = \frac{1}{d} \cdot \sum_{j\neq i}\pi_j^* \cdot \sum_{l=1}^n A_{li}A_{lj} \big[X_{lj}(1-X_{li}) - \E[X_{lj}(1-X_{li})] \big] \,.\\
\end{aligned}
\end{equation*}
For a fixed $l$, if we change $X_{li}$, the sum above changes by at most $\frac{1}{d} \cdot \sum_{j\neq i} \pi_j^* A_{li} A_{lj} $. On the other hand, for a fixed $l$ and $j\neq i$, if we change $X_{lj}$, the sum above changes by at most $\frac{1}{d} \cdot \pi_j^* A_{li}A_{lj} \leq \frac{1}{d}\cdot \lVert \pi^*\rVert_{\infty} A_{li}A_{lj}$. Note also that under the growing $m$ regime and by Cauchy-Schwarz inequality, we have
\begin{equation*}
\begin{aligned}
&\sum_{j\neq i} \frac{1}{d}\, \pi_j^* A_{li} A_{lj} \leq \lVert \pi^*\rVert_{\infty} \cdot \sum_{j\neq i} A_{lj} \cdot A_{li}A_{lj}  \\
&\leq \frac{1}{d}\, \lVert \pi^*\rVert_{\infty} \cdot \sqrt{\sum_{j\neq i} A_{lj} } \cdot \sqrt{\sum_{j\neq i} A_{li}A_{lj} }\\
&\leq \frac{1}{d}\,\lVert \pi^*\rVert_{\infty} \cdot \sqrt{3/2} \cdot \sqrt{mp} \cdot \sqrt{\sum_{j\neq i} A_{li}A_{lj} } \,.\\
\end{aligned}
\end{equation*}
Invoking Hoeffding's inequality, we have
\begin{equation*}
\begin{aligned}
&\Pr\bigg( \frac{1}{d}\,\sum_{j\neq i}\pi_j^* \cdot \sum_{l=1}^n A_{li}A_{lj} \big[X_{lj}(1-X_{li}) - \E[X_{lj}(1-X_{li})]\big] > t\,\bigg|\, \A^+  \bigg) \\
&\leq 2 \, \exp{- \frac{2t^2}{ \frac{1}{d^2}\lVert \pi^*\rVert^2_{\infty}\,\left(\sum_{l=1}^n (3/2)mp \sum_{j\neq i}A_{li}A_{lj} + \sum_{j\neq i}\sum_{l=1}^n A_{li}A_{lj}\right)  }}\\
&=\,2\,\exp{-\frac{2t^2}{  \frac{1}{d^2}\lVert \pi^*\rVert^2_{\infty} (3/2\,mp + 1) \left(\sum_{l=1}^n \sum_{j\neq i} A_{li}A_{lj} \right)   }}\\
&\leq 2\,\exp{-\frac{2d^2t^2}{ \lVert \pi^*\rVert^2_{\infty} (3/2\,mp + 1) (m-1) \frac{3}{2}np^2  }}\\
&\leq 2\,\exp{-\frac{2d^2t^2}{ \lVert \pi^*\rVert^2_{\infty} 9/4\,mp \cdot mnp^2  }}\\
&\left[\text{Setting $d = \frac{3}{2}mnp^2$}\right]\\
&= 2\,\exp{-\frac{2mnp^2 t^2}{ \lVert \pi^*\rVert^2_{\infty} mp }} = 2\,\exp{-\frac{2np t^2}{ \lVert \pi^*\rVert^2_{\infty}}} \,.
\end{aligned}
\end{equation*}
Similarly, we can bound the term $I_4^{(i)}$ using the same method of bounded difference. Specifically, we have
\begin{align*}
I_4^{(i)} &= \pi_i^* (M_{ii} - M_{ii}^* )\\
&= \pi_i^* \sum_{j\neq i} \sum_{l=1}^n \frac{1}{d}\, \left(A_{li} A_{lj} X_{li}(1-X_{lj}) - \E[X_{li}(1-X_{lj})]\right)\\
&\leq \lVert \pi^*\rVert_{\infty}\, \sum_{j\neq i} \sum_{l=1}^n \frac{1}{d}\, \left(A_{li} A_{lj} X_{li}(1-X_{lj}) - \E[X_{li}(1-X_{lj})]\right)\,.
\end{align*}
One can see that this can be bounded in the same way as $I_3^{(i)}$ shown above. We combine the probabilistic bound on $I^{(i)}_3$ and $I^{(i)}_4$ and apply union bound over all $i \in [m]$. We have
\begin{align*}
&\Pr\left(\exists \,i \in [m]\,:\, \abs{I_3^{(i)}} \lor \abs{I_4^{(i)}} > t \right) \leq 4 m \,\exp{-\frac{2npt^2}{\lVert\pi^*\rVert^2_{\infty}}}\\
&\leq 4 \,\exp{-\frac{2npt^2}{\lVert\pi^*\rVert^2_{\infty}} + \log m}\\
&\leq 4 \,\exp{-\frac{npt^2}{\lVert\pi^*\rVert^2_{\infty}}}\,,
\end{align*}
where the last inequality holds so long as $\frac{2npt^2}{\lVert\pi^*\rVert^2_{\infty}} > 2 \log m$. Pick $t = \frac{\sqrt{10}\sqrt{\log m}}{\sqrt{np}} \lVert\pi^*\rVert_{\infty}$. We have
$$ \abs{I_3^{(i)}} \lor \abs{I_4^{(i)}} < \frac{\sqrt{10}\sqrt{\log m}}{\sqrt{np}} \lVert\pi^*\rVert_{\infty} \,  $$
simultaneously for all $i \in [m]$ with probability at least $1 - 4\left( m^{-10} \right)$.
\end{proof}


~\\
We can obtain sharper bound for the term $I_1^{(i)}$ under the assumption that $mp \gtrsim \log n$ as well.
\begin{lemma}\label{lem:pi-ell2-bound-growing} Suppose that Condition $\A^+$ and Condition $\calE$ both hold. There exists a constant $C_2$ such that if $np^2 \geq \frac{e^{4\kappa}C_2}{\gamma^3} \left( \log np^2 \lor \log m\right) $ then
\begin{equation*}
\lVert \pi - \pi^* \rVert_2 \leq \frac{1}{\frac{\gamma}{6e^{2\kappa}}} \, \frac{C_6\sqrt{m}}{\sqrt{np}} \, \lVert \pi^* \rVert_{\infty}
\end{equation*}
with probability at least $1 - \exp{-12m}$ where $C_6$ is an absolute constant.
\end{lemma}

\begin{proof} We follow a similar strategy as in the proof of Lemma \ref{lem:pi-ell2-bound} by providing an upper bound on the RHS of (\ref{eqn:ssd-bound}). The denominator $\mu^*(M^*) - \norm{M - M^*}_2$ in (\ref{eqn:ssd-bound}) can be bounded in the same way as the proof of Lemma \ref{lem:pi-ell2-bound}, leading to 
$$ \mu(M^*) - \lVert M - M^* \rVert_{2} \geq \frac{\gamma}{6 e^{2\kappa}}. $$ 
The numerator of (\ref{eqn:ssd-bound}) can be bounded in the same way as the proof of Lemma A.11 of \citet{nguyen2022spectral} which improves by a factor of $\frac{1}{\sqrt{p}}$ over the upper bound on $\norm{{\pi^*}^\top (M - M^*)}$ obtained in Lemma \ref{lem:pi-ell2-bound}. Specifically, Lemma A.11 of \citet{nguyen2022spectral} states that conditioned on $\A^+$, with probability at least $1 - \ebr{-12m}$,
\begin{equation*}
\norm{{\pi^*}^\top (M - M^*)} \leq \frac{C_6\norm{\pi^*}_{\infty} \sqrt{m}}{\sqrt{np}}\,,
\end{equation*}
where $C_6$ is an absolute constant.
\end{proof}

\begin{lemma}\label{lem:I1-growing} Suppose that the setting and conclusion of Lemma \ref{lem:pi-ell2-bound-growing} hold. Then the following inequality holds for all $i \in [m]$ deterministically.
\begin{equation*}
\abs{I_1^{(i)}} \leq \frac{1 + \frac{\gamma}{12e^{2\kappa}}}{\frac{\gamma}{6e^{2\kappa}}} \, \frac{C_6}{\sqrt{np}} \, \lVert \pi^* \rVert_{\infty}\,.
\end{equation*}
\end{lemma}

\begin{proof} Using Cauchy-Schwarz inequality, we have
$$ I_1^{(i)} \leq \lVert \pi - \pi^* \rVert_2 \cdot \sqrt{\sum_{j\neq i} M_{ji}^2 } \,. $$
Following the same argument in the proof of Lemma \ref{lem:I1-const}, we have
\begin{align*}
 \sqrt{\sum_{j\neq i} M_{ji}^2 } \leq \left(1 + \frac{\gamma}{12e^{2\kappa}}\right) \,\frac{1}{\sqrt m} \,.
\end{align*}
Substituting the above bound and Lemma \ref{lem:pi-ell2-bound-growing} into (\ref{eqn:I1}) completes the proof.
\end{proof}

~\\
We are now ready to put all of the bounds on $I_1^{(i)}$, $I_3^{(i)}$, $I_4^{(i)}$ together.
\begin{theorem}\label{thm:pi-error-growing} Suppose that Condition $\A^+$ and Condition $\calE$ both hold. Suppose further that $np^2 > \frac{C^2_1}{\gamma} \left(\log np^2 \lor \log m\right) \lor \frac{e^{4\kappa}C_2}{\gamma^3} \left(\log np^2 \lor \log m\right) $ where $C_1, C_2$ are two absolute constants. Then
\begin{equation*}
\lVert \pi - \pi^* \rVert_{\infty} \leq \frac{1}{\frac{\gamma}{3}\left(1 - \frac{\gamma}{12e^{2\kappa}}\right)} \, \left( \,\frac{2C_5\sqrt{ \log m}}{\sqrt{np}}  +\frac{1 + \frac{\gamma}{12e^{2\kappa}}}{\frac{\gamma}{6e^{2\kappa}}} \, \frac{C_6}{\sqrt{np}}  \right) \, \lVert \pi^* \rVert_{\infty}
\end{equation*}
with probability at least $1 - 4 m^{-10} - \exp{-12m} $.
\end{theorem}

\begin{proof} The proof combines the conclusion of Lemma \ref{lem:I3-I4-growing} and Lemma \ref{lem:I1-growing}.
\end{proof}

\begin{reptheorem}{thm:entrywise-error} There exist absolute constants $C_0, C_1, C_2, C_7$ such that if $np^2 > C_0\log m \, \lor \frac{C^2_1}{\gamma} \left(\log np^2 \lor \log m\right) \lor \frac{e^{4\kappa}C_2}{\gamma^3} \left(\log np^2 \lor \log m\right) \lor \frac{C_7e^{6\kappa}}{\gamma^4}  \log m $ then
\begin{equation*}
\lVert \beta - \beta^* \rVert_{\infty} \leq \frac{4e^{2\kappa}}{\frac{\gamma}{3}\left(1 - \frac{\gamma}{12e^{2\kappa}}\right)} \, \left( \,\frac{2C_5\sqrt{\log m}}{\sqrt{np}}  +\frac{1 + \frac{\gamma}{12e^{2\kappa}}}{\frac{\gamma}{6e^{2\kappa}}} \, \frac{C_6}{\sqrt{np}}  \right)
\end{equation*}
with probability at least $1 - \exp{-\frac{np^2}{24}}  - \frac{2}{n^{10}} - 4m^{-10} - \exp{-12m} $.
\end{reptheorem}
\begin{proof} The proof is similar to that of Theorem \ref{thm:beta-err-const} except that we use the bound in Theorem \ref{thm:pi-error-growing} which is tighter than that in Theorem \ref{thm:pi-error-const}.
\end{proof}

\begin{reptheorem}{thm:top-K} Consider the setting of Theorem \ref{thm:entrywise-error}. Consider a top-$K$ estimator that first runs the spectral algorithm on the response data and returns the $K$ items with the highest parameter values. There exists a constant $C_K$ such that if $np \geq \frac{C_K\log m}{\Delta_K^2} $, then this top-$K$ estimator correctly identifies all of the top $K$ items
with probability at least $1- \bigO\left(m^{-10}\right) - \exp{-\bigO(np^2)} - \bigO\left(n^{-10}\right)$ where $C$ is an absolute constant.
\end{reptheorem}

\begin{proof} One can see that given sufficiently large $n$ such that 
$$ \lVert \beta - \beta^* \rVert_{\infty} < \Delta_K \,,$$
then the top $K$ items with the highest parameter values per the spectral algorithm is also the true top $K$ items.
\end{proof}

\newpage
\section{Proofs of Lower Bounds}

\subsection{Estimation Error Lower Bound}
In this section we prove the lower bound for the estimation error presented in the main paper. We emphasize that this lower bound is different from the Cramer Rao lower bound obtained in \citet{nguyen2022spectral}. Specifically, Cramer-Rao lower bounds apply only to the class of unbiased estimators. However, the information theoretic lower bound presented here applies to all statistical estimators, a strictly larger class of estimators. We first restate the theorem statement and defer the proof to later.

\begin{reptheorem}{thm:lower-bound-error} Consider the sampling model described in Section \ref{sect:problem} and further assume that $np = \Omega(1)$. There exists a class of Rasch models $\B$ such that for any statistical estimator, the minimax risk is lower bounded as
$$  \inf_{\hat \beta}\sup_{\theta^* \in \mathcal{T}, \beta^* \in \B}\, \E \lVert\hat \beta - \beta^* \rVert_2^2 = \Omega\left(\frac{m}{np}\right)\,. $$
\end{reptheorem}

~\\
The proof of Theorem \ref{thm:lower-bound-error}, which we defer to later, relies on the pairwise Fano's inequality \citep{thomas2006elements}.

\begin{lemma}\label{lem:fanos-minimax} (Pairwise Fano minimax lower bound) Suppose that we can construct a set $\calW = \{{\beta^*}^1, \ldots, {\beta^*}^M\}$ with cardinality $M$ such that 
$$ \max_{a\neq b\in[M]}\lVert \beta^{*a} - \beta^{*b} \rVert_2^2 \leq \delta^2 \quad\text{and}\quad \max_{a\neq b\in [M]}\kl(\Pr^a(X)\lVert \Pr^b(X)) \leq \xi\,, $$
where $\Pr^a(X)$ denotes the distribution over $X$ under a model parametrized by ${\beta^*}^a$. Then the minimax risk is lower bounded as
$$ \inf_{\hat\beta} \sup_{\beta^* \in \B} \E[\lVert \hat\beta - \beta^*\rVert_2^2] \geq \frac{\delta^2}{2}\bigg(1-\frac{\xi+\log 2}{\log M}\bigg)\,, $$
where $\hat\beta$ is the output of any statistical estimator.
\end{lemma}

~\\
Intuitively, the theorem says that if we can construct a set of models where every pair of model is sufficiently different (measured in $\ell_2^2$ distance) yet they parametrize `similar' distributions (measured in KL divergence), then any statistical estimator will fail to identify the correct model and thus suffer from a minimum estimation error. To construct the set of models, we follow the construction similar to that in the proof of Theorem 1 of \cite{shah2015estimation}. We first restate a coding theoretic due to \citet{shah2015estimation}.

\begin{lemma}\label{lem:packing-set-shah} (Lemma 7 of \citet{shah2015estimation}) For any $\alpha \in (0, \frac{1}{4})$, there exists a set of $M(\alpha) = \bigtheta(\exp{m})$ binary vectors $\{z^1, \ldots, z^{M(\alpha)}\} \subset \{0, 1\}^m $ such that
\begin{equation*}
\begin{aligned}
\alpha m \leq \lVert z^{a} - z^b \rVert_2^2 \leq m\quad &\forall a \neq b\in [1,\ldots, M(\alpha)] \,.\\
\end{aligned}
\end{equation*}
\end{lemma}

~\\
Consider a set $\{z^1, \ldots, z^{M(\alpha)}\} \subset \{0, 1\}^m $ of $m$-dimensional binary vectors given by as Lemma \ref{lem:packing-set-shah}. Let
$$ \beta^{*a} = \frac{\delta}{\sqrt m} \cdot z^a \quad \forall a\in [M(\alpha)]\,, $$
where $\delta$ is to be detemined later. It is easy to see that for $a\neq b \in [M(\alpha)]$
$$  \lVert \beta^{*a} - \beta^{*b} \rVert_2^2 = \frac{\delta^2}{m} \cdot \lVert z^{*a} - z^{*b} \rVert^2_2 \leq \delta^2 \,.$$

\begin{lemma}[Reverse Pinsker's inequality] Consider two probability measures $P$ and $Q$ defined on the same measure space $(\A, \calF)$ the KL divergence between $P$ and $Q$ can be bounded as
$$ \kl(P \lVert Q) \leq \left(\frac{\log e}{Q_{\min}}\right) \cdot \lVert P - Q \rVert_{TV}^2\,, $$
where $\lVert P - Q \rVert_{TV}$ is the total variation distance between the two distributions and $Q_{\min} := \min_{x \in \A} Q(x)$.
\end{lemma}

~\\
For our lower bound, it suffices to consider the case where $\theta_1 = \ldots = \theta_n = \frac{1}{2}$. Denote $\mathcal{T} = \{ \theta: \theta_1 = \ldots = \theta_n = \frac{1}{2} \} $.

\begin{lemma}\label{lem:kl-div-theta-known} Consider $\theta \in \mathcal{T}$ and any two alternative parameter sets $\beta^a$ and $\beta^b$. There exists a constant $c^* = c^*(\beta^*_{\min}, \beta_{\max}^*)$ such that
$$\kl(\Pr^a(X)\lVert \Pr^b(X)) \leq  \frac{e^2\log e}{2c^*} \cdot np\delta^2 \,. $$
\end{lemma}

\begin{proof}
As a short hand notation, define $\Pr_i^a := \frac{e^{\theta}}{e^{\theta} + e^{\beta^a_i}}$ and analogously for $\Pr_i^b$.
\begin{equation*}
\begin{aligned}
\kl(\Pr^a(X)\lVert \Pr^b(X)) &= \sum_{l=1}^n \sum_{i=1}^m \kl(\Pr^a(X_{li})\lVert \Pr^b(X_{li})) \\
&= np \sum_{i=1}^m \kl(\Pr^a \lVert \Pr^b)\\
&\text{[Using the inverse Pinsker's inequality]}\\
&\leq np \sum_{i=1}^m \frac{\log e}{c^*} \cdot 2(\Pr_i^a - \Pr^b_i)^2\\
&\leq \frac{np \,2\log e}{c^*} \cdot \sum_{i=1}^m  \frac{e^2 \delta^2}{4m}\\
&\leq \frac{e^2\log e}{2c^*} \cdot np\delta^2\,.
\end{aligned}
\end{equation*}
The second line follows from 
\begin{align*}
 \kl(\Pr^a(X_{li})\lVert \Pr^b(X_{li})) &= \Pr^a(X_{li}=*) \cdot \log \left(\frac{\Pr^a(X_{li}=*)}{\Pr^b(X_{li}=*)}\right) + \Pr^a(X_{li}=1) \cdot \log \left(\frac{\Pr^a(X_{li}=1)}{\Pr^b(X_{li}=1)}\right)\\
 & + \Pr^a(X_{li}=0) \cdot \log \left(\frac{\Pr^a(X_{li}=0)}{\Pr^b(X_{li}=0)}\right)\\
 &= (1- p) \cdot 0 + p \Pr^a_i \cdot \log\left(\frac{p \Pr^a_i}{p \Pr^b_i}\right) + p (1-\Pr^a_i) \cdot \log\left(\frac{p (1-\Pr^a_i)}{p (1-\Pr^b_i)}\right)\\
 &= p \cdot \left( \Pr^a_i \cdot \log\left(\frac{\Pr^a_i}{\Pr^b_i}\right) + (1-\Pr^a_i) \cdot \log\left(\frac{(1-\Pr^a_i)}{(1-\Pr^b_i)}\right)  \right)\\
 &= p \, \kl(\Pr^a \lVert \Pr^b)\,.
\end{align*} 
The second inequality comes from applying inverse Pinsker's inequality (for binary random variable) and the observation that
\begin{align*}
\min_{X_{li} \in \{0, 1\}} \Pr^b(X_{li}) &= \min\left\{ \frac{\sqrt e}{\sqrt{e} + e^{\beta^b_i}}, \frac{e^{\beta^b_i}}{\sqrt{e} + e^{\beta^b_i}} \right\}\\
&\geq \min\left\{ \frac{\sqrt e}{\sqrt{e} + e^{\beta^*_{\max}}}, \frac{e^{\beta^*_{\min}}}{\sqrt{e} + e^{\beta^*_{\min}}} \right\} =: c^* \,. \\
\end{align*}
The third inequality comes from
\begin{align*}
(\Pr_i^a - \Pr_i^b)^2 &= \left( \frac{e^{1/2}}{e^{1/2} + e^{\beta^a_i}} - \frac{e^{1/2}}{e^{1/2} + e^{\beta^b_i}}\right)^2\\
&= \left(\frac{\sqrt e \cdot(e^{\beta^a_i} - e^{\beta^b_i}) }{(\sqrt e + e^{\beta^a_i})(\sqrt e + e^{\beta^b_i})}   \right)^2\\
&\leq \left(\frac{\sqrt e (e^{\delta/\sqrt m} - 1) }{(\sqrt e + e^{\delta/\sqrt m})(\sqrt e +1)}   \right)^2\\
&\leq \left(\frac{e\delta}{2\sqrt m}\right)^2 = \frac{e^2 \delta^2}{4m}\,.
\end{align*}
The second last inequality comes from the observation that $\beta_i^a, \beta_i^b \in \{0, \frac{\delta}{\sqrt m}\}$. To see why the last inequality holds, one can verify the following useful inequality.
$$ \frac{e^x - 1}{(\sqrt e + e^x)(\sqrt e + 1)} \leq \frac{x}{2e} \,\forall x > 0 \,.$$
\end{proof}

~\\
Finally, the proof for Theorem \ref{thm:lower-bound-error} follows from Lemma \ref{lem:kl-div-theta-known} and Lemma \ref{lem:packing-set-shah}.
\begin{proof}(\textbf{Of Theorem \ref{thm:lower-bound-error}})
We can now apply Lemma \ref{lem:fanos-minimax}. The minimax risk is lower bounded as
$$  \inf_{\hat\beta} \sup_{\beta^* \in \B} \E[\lVert \hat\beta - \beta\rVert_2^2] \geq \frac{\delta^2}{2}\bigg(1-\frac{Cnp\delta^2 +\log 2}{m}\bigg)\,, $$
where $C$ is $\frac{e^2\log e}{2c^*}$.
Setting $\delta^2 = \frac{m}{2Cnp}$ gives us
$$ \inf_{\hat\beta} \sup_{\beta^* \in \B} \E[\lVert \hat\beta - \beta\rVert_2^2] \geq \frac{m}{4C\, np} \cdot \left(1 - \frac{m/2+\log 2}{m}\right) \gtrsim \frac{m}{np}\,.  $$
This completes the proof.
\end{proof}

\subsection{Top-$K$ Recovery Lower Bound}

We first restate a different version of Fano's inequality \cite{thomas2006elements} which will be useful to the construction of our lower bound. Note that this Fano's inequality is designed for hypothesis testing questions which is different from the pairwise Fano's inequality used to prove the lower bound for parameter estimation. An algorithm does not have to produce parameter estimates in order to produce top-$K$ estimate. Therefore, we need to develop a separate theorem proving the lower bound for top-$K$ recovery.

\begin{lemma} [Fano's inequality]\label{lem:top-k-fano} Consider a set of $L$ distributions $\{\Pr^1,\ldots, \Pr^L\}$. Suppose that we observe a random variable (or a set of random variables) $Y$ that was generated by first picking an index $A \in \{1,\ldots,L\}$ uniformly at random and then $Y \sim \Pr^A$. Fano's inequality states that any hypothesis test $\phi$ for this problem has an error probability lower bounded as
$$ \Pr[\phi(Y) \neq A] \geq 1 - \frac{\max_{a,b\in [L], a \neq b} \kl(\Pr^a(Y)\rVert \Pr^b(Y)) + \log 2 }{\log L}\,. $$
\end{lemma}
The intuition behind the above version of Fano's inequality is the same as that for the version used in proving the estimation error lower bound. Suppose that we can construct a set of models where the distributions parametrized by any pair of models are similar (as measured in terms of KL divergence). Then no statistical estimator can accurately identify a single model that has been uniformly chosen from this set.

~\\
\textbf{Construction of the models}. Let us consider the following constructions for $m-K+1$ models. For simplicity, let us consider the unnormalized parameter space for now. This is valid because we are only concerned about the KL divergence between any pair of models, which is not affected by the normalization step and only by the relative difference among the items. For model $a \in [K, K+1, \ldots, m]$,
$$ \beta^{*a}_{i} = \begin{cases} \delta &\text{if } i\leq K-1 \\ \delta &\text{if } i- K = a \\ 0 &\text{otherwise}  \end{cases} \,.$$
In other words, the $n-K+1$ models differ exactly by the identity of the $K$-th best item. It is also clear that $\Delta_K = \delta$ in this context. For top-$K$ recovery lower bound, it also suffices to consider the case where $\theta_1 = \ldots = \theta_n = \frac{1}{2}$.

\begin{lemma}\label{lem:kl-top-K} Consider the model construction described above. There exists a constant $c$ such that for any two models $a \neq b$, 
$$ \kl(\Pr^a \lVert \Pr^b) \leq c\, np\Delta_K^2 \,.$$
\end{lemma}

\begin{proof} We follow the same argument as used in the proof of Lemma \ref{lem:kl-div-theta-known}. We have
\begin{equation*}
\begin{aligned}
\kl(\Pr^a(X)\lVert \Pr^b(X)) &= \sum_{l=1}^n \sum_{i=1}^m \kl(\Pr^a(X_{li})\lVert \Pr^b(X_{li})) \\
&= np \sum_{i=1}^m \kl(\Pr^a \lVert \Pr^b)\\
&\text{[Using the inverse Pinsker's inequality]}\\
&\leq np \sum_{i=1}^m \frac{\log e}{c^*} \cdot 2(\Pr_i^a - \Pr^b_i)^2\\
&\leq \frac{np \,2\log e}{c^*} \cdot \sum_{i=1}^m  \frac{e^2 \delta^2}{4m}\\
&\leq \frac{e^2\log e}{2c^*} \cdot np\delta^2\,,
\end{aligned}
\end{equation*}
where $ c^* = \min\left\{ \frac{\sqrt e}{\sqrt{e} + e^{\beta^*_{\max}}}, \frac{e^{\beta^*_{\min}}}{\sqrt{e} + e^{\beta^*_{\min}}} \right\}$ and the constant $c$ in the lemma statement is $ \frac{e^2\log e}{2c^*} $.
\end{proof}

\begin{reptheorem}{thm:lower-bound-topK} Consider the sampling model described in Section \ref{sect:problem}. There is a class of Rasch model such that if 
$np \leq \frac{c_K\log m}{\Delta_K^2} \,,$
where $c_K$ is a constant, then any estimator will fail to identify all of the top $K$ items with probability at least $\frac{1}{2}$.
\end{reptheorem}

\begin{proof} Applying Lemma \ref{lem:top-k-fano} and Lemma \ref{lem:kl-top-K}, we have
\begin{align*}
\Pr[\phi(Y) \neq A] &\geq 1 - \frac{\max_{a,b\in [L], a \neq b} \kl(\Pr^a(Y)\rVert \Pr^b(Y)) + \log 2 }{\log L}\\
&\geq 1 - \frac{\frac{4np\delta^2}{c} + \log 2 }{\log (m-K+1)}\\
&\approx 1 - \frac{4np\delta^2}{c\log m}\,,
\end{align*}
where $c =  \frac{e^2\log e}{2c^*}$. The above lower bound tells us that if 
$$ np \leq \frac{c\log m}{8\Delta_K^2} $$
then $\Pr[\phi(Y) \neq A] \geq \frac{1}{2}$. This finishes the proof.
\end{proof}

\newpage
\section{Differentially Private Spectral Algorithm}

The following lemma establishes the desirable property of concentrated differential privacy that both post-processing and composition preserve concentrated differential-privacy. Intuitively, one can see that the construction of the private pairwise differential measurement in Equation (\ref{eqn:private-Yij}) is analogous to a data-preprocessing step. By ensuring that this step is private, the final output of the algorithm remains private as well. An analogous result for DP can be found in \cite{dwork2014algorithmic}.

\begin{lemma}[Lemma 2.3 of \citet{bun2016concentrated}]\label{lem:composition-cdp} Consider domains $\X$, $\Y$, $\Z$ and two functions $ f: \X \rightarrow \Y, f': \X\times \Y \rightarrow \Z$. Suppose that $f$ satisfies $\xi$-concentrated differential privacy and $f'$ satisfies $\xi'$-concentrated differential privacy. Define $f'' : \X\rightarrow \Y$ as $f''(x) = f'(x, f(x))$. Then $f''$ satisfies $(\xi + \xi')$-concentrated differential privacy.
\end{lemma}

~\\
The following lemma establishes the privacy guarantee of the discrete Gaussian mechanism.
\begin{lemma}[Theorem 4 of \cite{canonne2020discrete}]\label{lem:discrete-gaussian-guarantee} Let $\Delta, \epsilon > 0$. For an input domain $\X$, let $f: \X \rightarrow \bbZ$ satisfy $\lvert f(x) - f(x') \rvert \leq \Delta$ for all $x \simeq x'$ differing on a single entry. Define a randomized algorithm $M : \X \rightarrow \bbZ$ as $M(x) = q(x) + Y$ where $Y \sim \calN_{\bbZ}(0, \frac{\Delta^2}{\epsilon^2})$. Then $M$ satisfies $\frac{1}{2}\epsilon^2$-concentrated differential privacy.
\end{lemma}

The quantity $\Delta$ in the theorem statement is often referred to as the sensitivity of a query $f$. Simply put, it is the maximum change in output of a query $f$ when one changes the data of a single user.
With the above two Lemmas, we are ready to prove the privacy guarantee of the private spectral algorithm. This quantity will reappear in the analysis of the discrete Laplace mechanism.

\begin{replemma}{lem:privacy-guarantee} Consider a modified spectral algorithm that first adds $\calN_{\Z}\left(0, \frac{\sqrt{m(m-1)}}{{\epsilon^*}^2}\right)$-distributed noise to each pairwise differential measurement per Equation (\ref{eqn:private-Yij}). This algorithm then proceeds to use the private differential measurements in the same way as in the original algorithm. This modified spectral algorithm satisfies $\frac{1}{2}{{\epsilon^*}^2}$-concentrated differential privacy.
\end{replemma}

\begin{proof} Recall the construction of the pairwise differential measurement for a pair of items $i, j$ as
\begin{equation}
Y_{ij}(X) =  \sum_{l=1}^n A_{li} A_{lj} X_{li}(1-X_{lj})\,.
\end{equation}
It is clear that for any neighboring datasets $X\simeq X'$, $\lvert Y_{ij}(X) - Y_{ij}(X') \rvert \leq 1$ (sensitivity). Invoking Lemma \ref{lem:discrete-gaussian-guarantee}, we conclude that the privatized function
$$ \tilde Y_{ij}(X) = Y_{ij}(X) + Z_{ij} \,,$$
where $Z_{ij} \sim \calN_{\bbZ}\left(0, \frac{1}{\epsilon^2}\right) $ satisfies $\frac{1}{2}\epsilon^2$-concentrated differential privacy. Invoking lemma \ref{lem:composition-cdp} with $\Delta = 1$, we see that the privatized spectral algorithm that constructs $m(m-1)$ pairwise differential measurements where each non-private measurement is added with $\calN_{\bbZ}\left(0, \frac{1}{\epsilon^2}\right)$-distributed noise satisfies $\frac{m(m-1)}{2}\epsilon^2$-concentrated differential privacy. 

Using simple algebra, one can check that for $\epsilon = \frac{\epsilon^*}{\sqrt{m(m-1)}}$, $m(m-1)$ compositions of $\frac{1}{2}\epsilon^2$-concentrated differentially private queries (using $\calN_{\bbZ}\left(0, \frac{\sqrt{m(m-1)}}{\epsilon^2}\right)$-distributed noise for each query) satisfies $\frac{1}{2}{\epsilon^*}^2$-concentrated differential privacy overall. Now, since concentrated differential privacy is closed under post-processing, the entire private spectral algorithm also satisfies the same privacy guarantee as the private pairwise measurements computation step, which is $\frac{1}{2}{\epsilon^*}^2$-concentrated differentially private.
\end{proof}

\paragraph{Entrywise Error Analysis of the Private Spectral Algorithm.} During the submission process of this paper, an anonymous reviewer has kindly pointed out that our theoretical analysis would be more complete with a study of the entrywise error guarantee of the spectral algorithm. It is indeed possible to obtain some rough analysis as follows. Firstly, note that the spectral algorithm constructs a Markov chain. Per our analysis, the statistical noise for the pairwise transition probabilities can be bounded as 
$$ \lvert M_{ij} - M^*_{ij} \rvert = O\left( \frac{\sqrt{\log m}}{m\sqrt{np^2}}  \right)\, \quad\forall i\neq j \in [m]\,,$$
where $M_{ij}$ is the non-private estimate and $M_{ij}^*$ is the idealized estimate. On the other hand, the discrete Gaussian distribution `behaves' similarly to a continuous Gaussian distribution \cite{canonne2020discrete} (i.e., light-tailed). That is, with high probability, the amount of noise introduced by the Gaussian mechanism to the pairwise transition probabilities (cf. Equation (4)) is bounded as
$$ \left\lvert \frac{Z_{ij}}{d} \right\rvert = O\left( \frac{\sqrt{\log m}}{np^2\epsilon^*}  \right) \,. $$
Hence, the deviation of the \emph{private} pairwise transition probabilities from the idealized pairwise transition probabilities can be bounded as
$$ \lvert \tilde M_{ij} - M^*_{ij} \rvert = O\left(  \frac{\sqrt{\log m}}{m\sqrt{np^2}}  + \frac{\sqrt{\log m}}{np^2\epsilon^*} \right) \,.$$
One can see that given sufficiently large sample size, $np^2 = \Omega\left(\frac{m^2}{{\epsilon^*}^2}\right)$, the additive noise introduced by the Gaussian mechanism is on the same order as or dominated by the statistical noise and we obtain the same error guarantees as the non-private spectral algorithm (up to a constant factor). Formalizing this argument, one obtains the following entrywise error guarantee of the private spectral algorithm.

\begin{theorem} Consider the uniform sampling model described in Section \ref{sect:problem}. There exist constants $C_1, C_2$ such that if $np^2 \geq C_1' \left\{\log m, \frac{m^2}{(\epsilon^*)^2} \right\} $ and $mp \geq C_2 \log n$ then the output of the spectral algorithm satisfies
$$ \lVert \beta - \beta^* \rVert_{\infty} \leq \frac{C'\sqrt{\log m}}{\sqrt{np}} $$
with probability at least $1- \bigO\left(m^{-10}\right) -\bigO\left(n^{-10}\right)$ where $C'$ is an absolute constant.
\end{theorem}

\begin{proof} The proof is relatively simple and is thus omitted.
\end{proof}

\newpage
\section{Additional Experimental Details}

\textbf{Additional Experiments.} During the submission process, an anonymous reviewer has kindly suggested that we can perform additional experiments examining other aspects of the algorithm such as time complexity and how the performance varies with $p_0$ (the sampling probability when applying the sparse Gaussian mechanism). Firstly, in terms of time complexity, both randomized response and the Gaussian mechanism are essentially data processing algorithms and both very fast compared to underlying non-private algorithm. The time complexity for the Gaussian mechanism is $O(m^2)$ while that for randomized response is $O(mn)$ where $m$ is the number of items and $n$ the number of users. In many real life data sets, $n \gg m$, the Gaussian mechanism is a more efficient procedure. The effect of $p_0$ on the performance of the Gaussian mechanism depends on the privacy regime. The figure below shows how the performance of the sparse Discrete Gaussian mechanism varies with $p_0$. Note that the line in blue is the proposed sparse method in our paper with $p_0 = \frac{\log n}{n}$.
\begin{figure}[h]
\centering
\begin{subfigure}{.5\textwidth}
  \centering
  \includegraphics[width=1\linewidth]{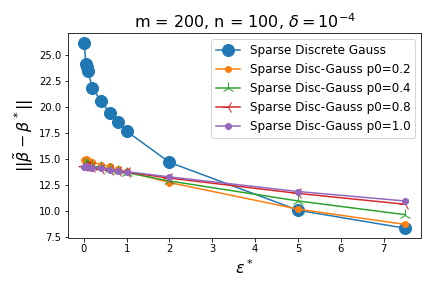}
\label{fig:sub1}
\end{subfigure}%
\caption{In the high-privacy regime (small $\epsilon^*$), a larger $p_0$ generally leads to better performance. In the low-privacy regime (larg $\epsilon^*$), using a sparse sampling graph results in more accurate estimates. In our work, we propose $p_0 = \frac{\log n}{n}$ as a basic starting point for designing a sparse Gaussian mechanism that performs decently well.}
\end{figure}
While the experiments in our work focus primarily on privacy mechanisms and parameter estimation under the Rasch model, we can also follow the experimental setup in previous works such as \cite{nguyen2022spectral} to explore private ranking applications. We also like to point out that a commonly studied algorithm in the ranking literature is spectral ranking \cite{negahban2017rank} also constructs a Markov chain in its procedure. The application of Gaussian mechanism in our algorithm can be extended to that setting to construct a differentially private algorithm specifically for ranking applications.

\textbf{Privacy Guarantee of Randomized Response with Shuffling.}
Firstly, we recall the procedure behind randomized response \cite{warner1965randomized} for binary responses. For each user $l$ and a response $X_{li}$ for $i \in [m]$. The following procedure ensures $(\epsilon, 0)$-differential privacy (cf. \citet{wang2016using}).
\begin{enumerate}
    \item The participant flips a coin with bias $\frac{e^\epsilon}{1+e^\epsilon}$ toward heads (outcome unknown to the statistician).
    \item If the outcome is heads, the participant answers truthfully. If the outcome is tails, the participant reports the answer opposite to the truth.
\end{enumerate}

We include here our implementation for randomized response.
\begin{python}
import numpy as np

def randomized_response(data, epsilon):
    # With probability e^epsilon/(1+e^epsilon), flip the label
    # For each user, we need to compute the number of responses, then divide epsilon by that number
    num_responses = np.sum(data != INVALID_RESPONSE, 0)    
    m, n = data.shape
    privatized_data = np.copy(data)
    
    for l in range(n):
        effective_epsilon = epsilon / num_responses[l] 
        # The epsilon we should be using for each user to flip their responses
        rand_p = np.random.rand(m) # Flip the binary response with probability 1/(1+e^epsilon)
        privatized_data[:, l] = np.where(np.logical_and(rand_p < 1/(1+np.exp(effective_epsilon)), \
                         data[:, l] != INVALID_RESPONSE), 1 - data[:, l], data[:, l])
    
    return privatized_data

\end{python}

Recall that the privacy guarantee of DP decays linearly with the number of queries, $m$ in our application. Hence, to ensure a $\epsilon^*$ privacy level overall for each user, each response $X_{li}$ by the same user $l$ needs to be randomized using the above procedure with $\epsilon = \epsilon^*/m$. Randomized response satisfies a special notion of DP known as local DP \citep{kasiviswanathan2011can}. It is also well known that local DP implies global DP. Furthermore, it has been shown recently that one can futher amplify the privacy guarantee of local DP mechanisms such as randomized response simply by permuting the randomized user data. In our application, since we are concerned primarily with estimating the item parameters, the algorithm is completely agnostic of this shuffling procedure.

\begin{lemma}[Theorem III.1 of \citet{feldman2022hiding}] For any domain $\calD$, let $\R^{(i)}: \Ss^{(1)}\times\ldots\times \Ss^{(i-1)} \times \calD \rightarrow \S^{(i)} $ for $i \in [n]$ be a sequence of algorithms such that $\R^{(i)}(z_{1:i-1},\cdot)$ is an $\epsilon_0$-DP local randomizer for all values pf auxiliary inputs $z_{1:i-1} \in \Ss^{(1)} \times \ldots \times \Ss^{(i-1)} $. Let $\A_s : \calD^{n} \rightarrow \Ss^{(1)} \times \ldots \times \Ss^{(n)} $ be the algorithm that given a dataset $x_{1:n} \in \calD^n $, samples a uniform random permutation $\sigma$ over $[n]$, then sequentially computes $z_i = \R^{(i)}(z_{1:i-1}, x_{\sigma(i)} )$ for $i \in [n]$ and outputs $z_{1:n}$. Then for any $\delta\in [0,1]$ such that $\epsilon_0 \leq \log\left(\frac{n}{16\log(2/\delta)}\right)$, $\A_s$ is $(\epsilon, \delta)$-DP, where
$$ \epsilon \leq \log \left( 1 + \frac{e^{\epsilon_0}-1}{e^{\epsilon_0}+1}\left( \frac{8\sqrt{e^{\epsilon_0}\log(4/\delta)}}{\sqrt{n}}  \right)  \right) \,.$$
\end{lemma}

Since the bound on $\epsilon$ in the theorem statement is a monotonous function in $\epsilon_0$, we can find $\epsilon_0$ using a simple bisection search algorithm. We include the code for computing $\epsilon_0$ below.

~\\
\begin{python}
def find_effective_epsilon0_rr(overall_epsilon, overall_delta, n):
    # epsilon0 is the maximum epsilon0 such that
    # e^eps = 1 + (e^eps0 - 1)/(e^eps0+1) (8 sqrt(e^eps0 log (4/delta))/sqrt(n) + 8e^eps/n) 
    # Solve for eps0

    def eps_estimate(eps0):
        eeps0 = np.exp(eps0)
        temp1 = (8 * np.sqrt(eeps0 * np.log(4/overall_delta)))/(np.sqrt(n))
        temp2 =  8 * eeps0/n
        return np.log(1 + (eeps0-1)/(eeps0+1) * (temp1 + temp2))
    
    # Just do bi-section search
    lower = 0.00001
    upper = np.log(n/(16*np.log(2/overall_delta)))
    
    if upper < 0:
        return overall_epsilon
    
    while eps_estimate(lower) > overall_epsilon:
        lower /= 2
    
    while upper - lower > 1e-8:
        mid = (upper + lower)/2
        if eps_estimate(mid) < overall_epsilon:
            lower = mid
        else:
            upper = mid
    
    return mid
\end{python}

~\\
\textbf{The discrete Laplace Mechanism.} The discrete Laplace mechanism is another noise-adding mechanism to ensure privacy that we consider in this work. Firstly, we restate the definition of the discrete Laplace mechanism \citep{ghosh2009universally}.

\begin{definition} Let $t > 0$. The discrete Laplace distribution with scale parameter $t$ is denoted $\text{Lap}_{\bbZ}(t)$. It is a probability distribution supported on the integers and defined by 
$$ \underset{X\leftarrow \text{Lap}_{\bbZ}(t) }{\Pr}[X= x] = \frac{e^{1/t}-1}{e^{1/t}+1} \cdot e^{-\lvert x\rvert/t}$$ for all $x\in \bbZ$.
\end{definition}
The Laplace mechanism adds discrete noise to a non-private query, with variance proportional to the sensitivity of the query.
\begin{lemma}[Theorem 1 of \cite{ghosh2009universally}]\label{lem:discrete-laplace-guarantee} Let $\Delta, \epsilon > 0$. For an input domain $\X$, let $f: \X \rightarrow \bbZ$ satisfy $\lvert f(x) - f(x') \rvert \leq \Delta$ for all $x \simeq x'$ differing on a single entry. Define a randomized algorithm $M : \X \rightarrow \bbZ$ as $M(x) = q(x) + Y$ where $Y \sim \text{Lap}_{\bbZ}(0, \frac{\Delta}{\epsilon})$. Then $M$ satisfies $\epsilon$-differential privacy.
\end{lemma}
It is possible to draw the discrete Laplace distribution using the following code.

~\\
\begin{python}
import numpy as np 

def laplace_noise(eps):
    def zero_prob(alpha):
        return 0 if np.random.rand() < (1.0 - alpha)/(1.0 + alpha) else 1

    def sign_prob():
        return -1 np.random.rand() < 0.5 else 1

    alpha = np.exp(-eps) # Between 0 and 1
    # This implementation is based on the usual geometric distribution
    # With probability (1-alpha)/(1+alpha), return 0
    # Otherwise, draw from the geometric distribution but with random sign flipping

    return zero_prob(alpha) * np.random.geometric(1-alpha) * sign_prob()

\end{python}


~\\
\textbf{Drawing from the discrete Gaussian distribution.} The following code can be used to draw from the discrete Gaussian distribution. This is based on Algorithm 1 in \cite{canonne2020discrete}.

\begin{python}
import numpy as np

def exponential_bernoulli_sample(gamma):
    assert(gamma >= 0)
    if 0 <= gamma <= 1:
        K = 1
        while(1):
            A = 1 if np.random.rand() < gamma/K else 0
            if A == 0:
                break
            else:
                K += 1
        return 1 if K % 2 == 1 else 0
    else:
        for k in range(1, int(np.floor(gamma) + 1)):
            B = exponential_bernoulli_sample(1)
            if B == 0:
                return 0
        return exponential_bernoulli_sample(gamma - np.floor(gamma))

def discrete_gaussian_sample(sigma):
    t = int(np.floor(sigma) + 1)
    
    while(1):
        U = np.random.choice(t)
        D = exponential_bernoulli_sample(U/t)
        if D == 0:
            # Restart
            return discrete_gaussian_sample(sigma)
        
        V = 0
        while(1):
            A = exponential_bernoulli_sample(1)
            if A == 0:
                break
            V += 1

        B = 1 if np.random.rand() < 1./2 else 0
        if B == 1 and U == 0 and V == 0:
            return discrete_gaussian_sample(sigma)
        Z = (1 - 2*B) * (U + t * V)
        C = exponential_bernoulli_sample((np.abs(Z)-sigma**2/t)**2/(2*sigma**2))
        
        if C == 0:
            return discrete_gaussian_sample(sigma)
        return Z

\end{python}

~\\
\textbf{The private spectral algorithm.} We now have all the necessary components to construct the private spectral algorithm. We include the implementation of the private spectral algorithm including the function to subsample for a random graph in the construction of the sparse Markov chain. For the code to run and produce our experiment results, refer to the attached python script in the supplementary materials.

\begin{python}

def subsampl_graph(m, p):
    subsample_graph = np.zeros((m, m))
    
    for i in range(m-1):
        for j in range(i+1, m):
            if np.random.rand() < p:
                subsample_graph[i, j] = 1
                subsample_graph[j, i] = 1
    
    return subsample_graph


def construct_markov_chain_private(performances, lambd=1., epsilon=0.1, subsample_graph=None):
    m = len(performances)
    
    if subsample_graph is None:
        subsample_graph = np.ones((m, m))
    
    D = np.ma.masked_where(performances == INVALID_RESPONSE, performances)
    D_compl = 1. - D
    M = np.ma.dot(D, D_compl.T)
    
    A = np.ma.masked_where(performances == INVALID_RESPONSE, np.ones_like(performances))
    B = np.ma.dot(A, A.T)
    
    np.fill_diagonal(M, 0)
    np.nan_to_num(M, False)
    M = np.round(M)
    
    M_non_priv = np.copy(M)
    M_add = np.copy(M)
    
    # Add discrete Gaussian noise
    for i in range(m):
        for j in range(m):
            if j != i and M[i, j] != 0 and np.abs(subsample_graph[i, j] - 1) < 1e-6:
                noise = discrete_gaussian_sample((1./epsilon))
                M[i, j] = max(1, M[i, j] + noise)    
                M_add[i, j] = noise
                
            if j != i and np.abs(subsample_graph[i, j] - 0) < 1e-6:
                M_non_priv[i, j] = 0
                M[i, j] = 0
                M_add[i, j] = 0
                    
    # Add regularization to the 'missing' entries
    M = np.where(np.logical_or((M != 0), (M.T != 0)), M+lambd, M)
    
    # d = np.ma.sum(M, 1) + 1
    d = np.ones((m,)) * np.ma.max(np.ma.sum(M, 1) + 1)

    for i in range(m):
        di = d[i]
        M[i, :] /= di
        M[i, i] = 1. - np.sum(M[i, :])

    return M_non_priv, M_add, M, d



def spectral_estimate_private(A, lambd=1., epsilon=0.1, subsample_graph=None, 
                                max_iters=1000, eps=1e-5):
    _, _, M, d = construct_markov_chain_private(A, lambd=lambd, epsilon=epsilon,
                                                subsample_graph=subsample_graph)
    assert(not np.any(np.isnan(M)))
    from scipy.sparse import csc_matrix

    M = csc_matrix(M)
    m = len(A)
    
    pi = np.ones((m,)).T
    for _ in range(max_iters):
        pi_next = (pi @ M)
        pi_next /= np.sum(pi_next)
        if np.linalg.norm(pi_next - pi) < eps:
            pi = pi_next
            break
        pi = pi_next
        
    pi = pi.T
    # pi = np.maximum(pi, 1e-12)
    pi /= np.sum(pi)
    pi = (pi/d)/np.sum(pi/d)
    assert(not np.any(np.isnan(pi)))
    beta = np.log(pi)
    beta = beta - np.mean(beta)
    assert(not np.any(np.isnan(beta)))
    return beta

\end{python}

~\\
\textbf{Converting Concentrated Differential Privacy to Approximate Differential Privacy.}

\begin{lemma}[Lemma 3.5 of \cite{bun2016concentrated}] Let $M$ be a $\rho$-concentrated differentially private algorithm. Then $M$ also satisfies $(\epsilon,\delta)$-approximate differential privacy for all $\delta > 0$ and 
$$ \epsilon = \rho + \sqrt{4\rho \log(1/\delta)} \,.$$
Generally, to achieve a desired $(\epsilon,\delta)$-differential privacy guarantee, it suffices to satisfy $\rho$-CDP with 
$$ \rho \approx \frac{\epsilon^2}{4\log(1/\delta)} \,. $$
\end{lemma}

Recall that the discrete Gaussian mechanism that adds $\calN_{\bbZ}\left(0, \frac{1}{\epsilon_0^2}\right)$ noise to a non-private query satisfies $\frac{1}{2}\epsilon_0^2$-CDP. By setting $\frac{1}{2}\epsilon_0^2$ to the approximation in the theorem statement above, we can solve for $\epsilon_0$ using the following one-line code.

\begin{python}
import numpy as np

def find_effective_epsilon0_zgauss(overall_epsilon, overall_delta):
    return overall_epsilon/np.sqrt(2 * np.log(1./overall_delta))

\end{python}

\bibliography{reference}